\documentclass[12pt,a4paper]{article}
\usepackage[utf8]{inputenc}
\usepackage[T1]{fontenc}
\usepackage{cite}
\usepackage{amsmath,amssymb,amsfonts}
\usepackage{algorithmic}
\usepackage{graphicx,color}
\usepackage{xurl}
\usepackage{changepage}
\usepackage{marvosym}
\usepackage{textcomp}
\usepackage{hyperref}
\usepackage{subcaption}
\usepackage{nameref}
\usepackage{float}
\usepackage[table]{xcolor}
\usepackage{array}
\usepackage{soul}
\usepackage{multirow}
\usepackage{booktabs}
\usepackage[margin=1in]{geometry}
\usepackage{setspace}
\usepackage{titlesec}
\usepackage[section]{placeins}
\usepackage{longtable}
\usepackage{tabularx}
\usepackage{multirow}
\usepackage{array}
\usepackage{adjustbox}

\titleformat{\section}{\normalfont\Large\bfseries}{\thesection}{1em}{}
\titleformat{\subsection}{\normalfont\large\bfseries}{\thesubsection}{1em}{}

\begin{document}

\title{\textbf{Remote Sensing Image Classification Using Deep Ensemble Learning}}

\author{
Niful Islam\textsuperscript{1}, Md. Rayhan Ahmed\textsuperscript{2}, Nur Mohammad Fahad\textsuperscript{3}, \\
Salekul Islam\textsuperscript{4}, A.K.M. Muzahidul Islam\textsuperscript{3}, \\
Saddam Mukta\textsuperscript{5}, Swakkhar Shatabda\textsuperscript{6} \\
\\
\small{\textsuperscript{1}Department of Computer Science and Engineering, Oakland University, Michigan, USA} \\
\small{\textsuperscript{2}Irving K. Barber Faculty of Science, University of British Columbia, Kelowna, Canada} \\
\small{\textsuperscript{3}Department of Computer Science and Engineering, United International University, Dhaka, Bangladesh} \\
\small{\textsuperscript{4}Department of Electrical and Computer Engineering, North South University, Dhaka, Bangladesh} \\
\small{\textsuperscript{5}Department of Software Engineering, LUT School of Engineering Sciences, LUT University, Finland} \\
\small{\textsuperscript{6}Department of Computer Science and Engineering, BRAC University, Dhaka, Bangladesh} \\
\\
\small{Corresponding Author: Saddam Mukta (e-mail: saddam.mukta@lut.fi)}
}

\date{}

\maketitle

\begin{abstract}
Remote sensing imagery plays a crucial role in many applications and requires accurate computerized classification techniques. Reliable classification is essential for transforming raw imagery into structured and usable information. While Convolutional Neural Networks (CNNs) are mostly used for image classification, they excel at local feature extraction, but struggle to capture global contextual information. Vision Transformers (ViTs) address this limitation through self attention mechanisms that model long-range dependencies. Integrating CNNs and ViTs, therefore, leads to better performance than standalone architectures. However, the use of additional CNN and ViT components does not lead to further performance improvement and instead introduces a bottleneck caused by redundant feature representations. In this research, we propose a fusion model that combines the strengths of CNNs and ViTs for remote sensing image classification. To overcome the performance bottleneck, the proposed approach trains four independent fusion models that integrate CNN and ViT backbones and combine their outputs at the final prediction stage through ensembling. The proposed method achieves accuracy rates of 98.10 percent, 94.46 percent, and 95.45 percent on the UC Merced, RSSCN7, and MSRSI datasets, respectively. These results outperform competing architectures and highlight the effectiveness of the proposed solution, particularly due to its efficient use of computational resources during training.
\end{abstract}


\vspace{1em}

\section{Introduction}
\label{sec:introduction}
Remote sensing (RS) refers to the technique of accumulating information about an entity from a certain distance, commonly using satellite, unmanned aerial vehicle (UAV), or other platforms \cite{boulila2021rs}. Remote sensing imagery provides a unique opportunity for retrieving information through the process of image interpretation and classification since it covers a large geographic area with high temporal frequency. The increasing volume of RS data has opened new studies and research opportunities in the fields of natural resource exploration, environment management, earthquake prediction, urban planning, and many more \cite{bazi2021vision}. All these sources of RS images generate a need for an automatic process for knowledge mining. Using image classification techniques, this vast amount of images can be organized according to their context \cite{mehmood2022remote}.\par 
Among popular image classification techniques, deep learning is the most popular one. Convolutional Neural Networks (CNNs) are the most common deep learning architectures for image recognition tasks. These deep learning models are specifically designed to process and extract features from image data, making them immensely effective in tasks such as object detection, image classification, and facial recognition \cite{maracani2023domain}. CNNs are composed of layers that learn and represent visual patterns collectively. These layers include fully connected, pooling, and convolutional ones. One of the most difficult challenges in training CNNs is the need for large labeled datasets \cite{gupta2023novel}. Transfer Learning, however, has become a potent method to get around this restriction and boost the effectiveness of training CNNs. Transfer learning entails using pre-trained models that have been optimized on smaller, task-specific datasets after being trained on massive datasets like ImageNet. Transfer learning enables CNNs to take advantage of the pre-trained model's representations and generalized patterns, which are shared by many different types of images \cite{bisheh2023image}. This knowledge transfer enables the CNN to learn more effectively on the target dataset, even when the labeled data is limited. Transfer learning also aids in combating overfitting because the trained model has already acquired useful representations from a large body of data \cite{kumar2023deep}. The initial layers of the pre-trained CNN's weights are frozen during the transfer learning process, and only the top layers are retrained using the target dataset. By using this method, the model is able to retain the low-level features it has already learned while adjusting to the specific features found in the new dataset. \par 
One of the biggest limitations of CNNs is that its limited ability to capture global context \cite{he2023hctnet}. Vision transformers (ViT) came as a solution to this problem \cite{dosovitskiy2020image}. The ViT introduces a new paradigm for image understanding, inspired by the success of transformers in natural language processing \cite{vaswani2017attention}. This architecture gets around CNN limitations by capturing the distant dependencies and taking into account an image's overall context. ViT works by dividing an image into smaller patches, which are then linearly projected by multiplying with a learnable weight matrix. These tokens are then fed into a typical transformer architecture made up of feed-forward neural networks and self-attention layers. The model gains the ability to focus on pertinent patches and draw out valuable representations from the image during training. The learned representations can then be used for a variety of downstream tasks such as image classification, object detection, and segmentation. The ViT has displayed astounding performance on a variety of benchmarks, competing with or even outperforming CNNs in some tasks. On massive datasets like ImageNet, it has produced state-of-the-art results in image classification, demonstrating its potential as a potent substitute for conventional CNN architectures. Following the outstanding performance of ViT, many vision transformer bases architectures have also emerged, such as data efficient image transformer (DeiT) \cite{touvron2021training}, swin transformers \cite{liu2021swin}, and others. \par 
Since the feature extraction mechanisms of CNNs and Vision Transformers are fundamentally different, their integration allows a model to exploit the complementary strengths of both architectures and achieve strong image recognition performance \cite{jiang2023marine}. Although hybrid models that combine multiple feature extractors often outperform standalone architectures, our study reveals a performance bottleneck in this approach. Experimental analysis shows that adding more CNN and ViT components results in insignificant performance improvements while increasing the computational cost, mainly due to overlapping feature representations that limit the contribution of additional extractors. To overcome this limitation, we propose an ensemble strategy based on a soft voting mechanism. In this work, we present a fusion architecture that integrates CNN and ViT models for remote sensing image classification, where four distinct fusion models composed of CNN and ViT backbones are trained independently and later ensembled through soft voting to produce the final prediction. The proposed method has been evaluated on the UC Merced, RSSCN7, and MSRSI datasets, and the results demonstrate outstanding classification performance across all datasets. In addition, we compare the proposed approach with several existing architectures and provide a detailed discussion that explains the reasons for its superior classification performance. In conclusion, the main contributions of this work are summarized as follows.
\begin{itemize}
    \item We have presented a novel architecture for classifying remote sensing images that is comprised of vision transformers and convolutional neural networks (CNNs). The architecture addresses the performance bottleneck issue by integrating a soft voting mechanism.
    \item The proposed model is tested on three benchmark datasets of remote sensing images, and the model performs significantly well in classifying different types of images.
    \item We have presented a comparison with different models and made a conclusive discussion about their performance. 
\end{itemize}
The rest of the article is organized as follows. The summary of the related works in the field of remote sensing images is presented in Section \ref{sec:rw}. Section \ref{sec:method} contains the detailed description of the proposed method, followed by the results in Section \ref{sec:results}. The article terminates in Section \ref{sec:conclusion}.

\section{Related Work}
\label{sec:rw}
Aerial classification from the remote sensing data has gained a lot of attention due to the comprehensive exploration of the locations and environmental monitoring. The advancement of AI can provide accurate classification of scenarios from remote sensing images, analyze the features of those scenarios, and improve the effectiveness and scalability of remote sensing applications. \par 
In recent studies, a combination of DL and ML models has been widely used for segmentation, localization, or classification purposes. Shen et al. \cite{shen2022attention} developed a dual-model deep features fusion technique for classifying the remote sensing images from 4 different datasets. They utilized two CNNs for feature extraction. The features were then passed through a cascade global–local network (ACGLNet) that filtered out repeated background information. The proposed method obtained a satisfactory accuracy of 98.14\% in UCM, 94.44\% in AID, 99.50\% in PatternNet, and 96.02\% in the OPTIMAL-31 dataset, respectively. Hilal et al. \cite{hilal2022deep} proposed a new deep transfer learning-based fusion model named DTLF-ERSIC. It extracted features from Discrete Local Binary Pattern (DLBP), ResNet, and EffecientNet which were merged later. This strategy was found to provide better performance compared to the base classifiers. Huang et al. \cite{huang2023semi}  combined supervised (SA) and unsupervised alignment (UA) techniques to propose a new bidirectional sample alignment technique for remote sensing scene classification. The supervised approach completes the feature alignment between source and target samples using the label of the samples, whereas the unsupervised approach aims to align the global features without demanding class information. They conducted these experiments on several datasets and obtained satisfactory performance. Ghadi et al. \cite{ghadi2022robust} performed object categorization and scene classification on remote sensing data using feature fusion techniques. They implemented a hybrid model employing fuzzy c-means, CNN, and FCN in order to segment and classify the images. The evaluation of the proposed techniques reports a mean accuracy of 97.73\% on the AID dataset, 98.75\% on the UCM, and 96.57\% on the RESIEC dataset, respectively. Alem and Kumar \cite{alem2022deep} developed a land cover and land use classification system using the deep learning approaches over the remote sensing images. They utilized CNN to interpret the images as well as extract the features, and developed a fine-tuned transfer learning model for classification. This model is evaluated on the UCM dataset and obtains an accuracy of 88.10\%. Al-Jabbar et al. \cite{al2023ebola} developed a modified Darknet-53 architecture using ebola optimization for scene classification and enhancing the security of smart cities. The Darknet-53 architecture was utilized to extract the feature from the remote sensing images of RSSCN7 dataset. Ebola optimization was used to optimize the hyperparameter and finally, graph convolution was used to perform the classification task. The proposed method attains a sublime accuracy of 99.52\%. Thirumaladevi et al. \cite{thirumaladevi2023remote} experimented with transfer learning with two state-of-the-art CNN models (AlexNet and VGG). The model extracted features with pretrained CNN feature extractor which are then passed to the SVM for final classification. Out of the two CNN models compared, VGG16 was found to produce better results. Vinaykumar et al. \cite{vinaykumar2023optimal} introduced a new method named optimal guidance-whale optimization algorithm (OG-WOA) for selecting relevant features in order to reduce the overfilling problem. They extracted features from AlexNet and ResNet which are then passed through OG-WOA where irrelevant features are deleted. Finally, the refined features make the classification through a shallow BiLSTM network. The proposed framework produced an excellent classification performance in classifying remote sensing images. Wang et al. \cite{wang2022remote} proposed a custom model made of residual dense attention block (RDAB) and coordinate attention (CA) module. RDAB module integrates dense connections as well as skip connections, leveraging the advantages of DenseNets and ResNets. Wang et al. \cite{wang12022remote} presented a novel two-stage based CNN architecture that consists of MS-Res block and Split Attention block. MS-Res block is made up by modifying the inception block with a hierarchical residual connection. The Split Attention module is integrated for cross-channel feature mapping. These two blocks are combined parallelly to construct the final model. Wang et al. \cite{wang2023large} introduced a model named Large kernel Sparse ConvNet weighted by Multi-frequency Attention (LSCNet) for classifying remote sensing images. LSCNet effectively addresses the need for larger receptive fields by utilizing two parallel convolutional layers. These layers extract features simultaneously and then combine them, eliminating the need for a large convolutional layer which reduces resource consumption. Additionally, the integration of an adaptive sparse optimization strategy and multi-frequency attention module enhanced performance and optimized the network. Although the proposed model is relatively small at 10.69 MB, it required 500 epochs to train due to the absence of transfer learning. Wang et al. \cite{wang2023remote} developed a three-steam architecture named Multi-stage Self-Guided Separation Network (MGSNet). This network consists of three branches: the background branch, the main branch, and the target branch. The authors extracted the target and the background representations with adaptive threshold segmentation which are passed to the background and target branches, respectively. The main branch, on the other hand, received the original image. The three streams extracted features simultaneously which are fused before classification. Since each stream consists of only three convolution blocks, the model is extremely lightweight. Zhang et al. \cite{zhang2023morphological} proposed a newly developed architecture named spatial-logical aggregation network (SLA-NET) for classifying hyperspectral images. This network consists of two key modules (i.e., the feature extractor and the adaptive feature fusion). The feature extractor consists of the spatial extractor stream and the morphological extractor stream. The two streams extract features simultaneously, which are fused with the adaptive feature fusion mechanism.

Recently, computer vision transformers have emerged as a powerful tool that outperforms the classical CNNs in the majority of the image classification datasets \cite{dosovitskiy2020image}. Therefore, it is being extensively applied for recognizing remote sensing images. Chaib et al. \cite{chaib2022co} extracted features from four ViT models and merged the feature vectors to obtain a raw dataset. Since employing multiple models can result in redundant features, they applied feature selection technique for data scrubbing. Finally, the classification was done through support vector machine (SVM). Swin Transformer is another recent addition in the computer vision transformer family that incorporates a shifted window and hierarchical feature fusion for image recognition \cite{liu2021swin}. Hao et al. \cite{hao2022two} presented a two-stream Swin Transformer network for RS image classification. In the first stream (original stream), the image is directly passed through the Swin Transformer, and the second stream (edge stream) passes the image through an edge detector made of sobel operator before passing it to the Swin Transformer. The seams are fused afterward to construct a robust classifier. Xu et al. \cite{xu2022vision} proposed a knowledge distillation based architecture named ET-GSNet. In this model, the vision transformer works as the teacher model and ResNet18 as the student model. They evaluated ET-GSNet on AID, NWPU-RESISC45, UCM, and  OPTIMAL-31 datasets. 

 While both CNNs and vision transformers have their own limitations, researchers explored combining both methods for effective classification. Roy et al.\cite{roy2023multimodal} fused hyperspectral and LiDAR data for remote sensing classification. They extracted CNN features, tokenized them into patch tokens, and fused these with a LiDAR-derived CLS token using a multihead cross-patch attention mechanism within a transformer encoder. Xu et al. \cite{xu2024bridging} merged CNN and transformer and designed a dual-branch network where a Dynamic-CNN branch captures local multiscale features and a Gaussian Transformer branch captures global dependencies, which are interactively fused via a cross-attention module. Similarly, Wang et al. \cite{wang2025tgf} combined transformer and gist CNN techniques to design a fusion network for multimodal remote sensing data. Their architecture began with a feature reconstruction module that decomposed hyperspectral and SAR (Synthetic Aperture Radar) inputs to reduce redundancy. Then, a transformer-based spectral module extracted global channel dependencies, while a gist CNN-based spatial module captured fine-grained spatial details. Finally, these distinct features were fused through a cross-attention fusion mechanism. Since the fusion models benefit from both CNNs and transformers, they produce better accuracy than single-stream architectures. 

Fu et al. \cite{fu2021example} utilized a graph convolutional network for semi-supervised remote sensing image classification. They introduced example-feature graph techniques for preserving the local geometrical distribution of the data. This experiment was evaluated on four public datasets and demonstrated a satisfactory performance. Yu et al. \cite{yu2019attention} developed an attention-based mechanism to perform the aerial scene classification. They introduced the attention GAN to address the scarcity of training data. After augmentation, a context-aggregation-based feature fusion architecture is applied to obtain the contextual features from the images of four publicly available datasets. Finally, the loss is calculated to yield the model’s performance. Yang et al. \cite{yang2022remote} designed a new method to evaluate the quality of remote sensing images using the node entropy calculation. The node module selection selects the images node from the labeled data, calculates the node entropy between the node images and unlabeled images, and locates the same distance node images in the same boundary. This study utilized this technique on AID and RSSCN7 datasets and obtained the accuracy of 92.48\% and 90.37\%, respectively. An efficient way to evaluate the proposed ViTs and CNNs combination is by comparing it with the existing ensemble methods that use mixed backbones. For example, several researchers have shown that ensemble methods that integrate CNNs with better ViT models are beneficial for various applications \cite{kyathanahally2022ensembles, wei2022ensemble, ashraf2023synthensemble, wu2021cvt}. A study achieved remarkable results in biodiversity image classification by using ensembles of Data-Efficient Image Transformers (DeiTs), in which a CNN was utilized as a teacher model for a ViT student \cite{kyathanahally2022ensembles}. To diagnose bronchial asthma, a hybrid ensemble model that combines CNNs and ViTs was also created. It uses a hierarchical framework that captures features throughout receptive fields \cite{wei2022ensemble}. Additionally, a fusion model that integrated CNNs and ViTs was presented in a recent work for the multi-label chest X-ray image classification \cite{ashraf2023synthensemble}. Another study reports increased accuracy and decreased error rates by employing a CvT model, obtaining reliable accuracy \cite{wu2021cvt}. Our proposed ensemble approach is aligned with this research trend and highlights the advantages of combining model fusion with a global context, capturing an effective approach to improve the efficacy and robustness of image classification tasks.

    \begin{longtable}{|p{1.5cm}|p{1.9cm}|p{2.5cm}|p{3.9cm}|p{3.5cm}|}
\caption{Comparison with existing works.}
\label{tab:compare-lit} \\

\hline
\textbf{Paper}  & \textbf{Dataset} & \textbf{Model} & \textbf{Drawbacks} & \textbf{Our Solution} \\
\hline
\endfirsthead

\hline
\textbf{Paper} & \textbf{Dataset} & \textbf{Model} & \textbf{Drawbacks} & \textbf{Our Solution} \\
\hline
\endhead

\hline
\endfoot

\hline
\endlastfoot
        Yu et al. \cite{yu2019attention} & UCM, AID, RSSCN7 & Attention GAN & No comparison with other transfer learning models, lower classification accuracy. & An ablation study comparing other approaches is present. Has higher classification accuracy.  \\ \hline
        Shen et al. \cite{shen2022attention}& UCM, AID, PatternNet, OPTIMAL-31 & ResNet50, DenseNet121, Attention Cascade Net & No transfer learning employed. Therefore, has a large number of trainable parameters. & We integrated pretrained models. Therefore, total trainable parameters are slightly above 8 million. \\ \hline
        Ghadi et al. \cite{ghadi2022robust} & AID, UCM, ResisC45 & fuzzy c- means, CNN, Feature fusion. & Incorporates machine learning algorithms that cannot be sped up by GPU. Therefore, very inefficient to train & The model can be sped up by GPU. \\ \hline
        Alem and Kumar \cite{alem2022deep} & UCM & Custom CNN model & Has low classification accuracy. & Achieves higher classification accuracy. \\ \hline
        Wang et al. \cite{wang2022remote} & UCM, AID & Multiple-Instance Learning with a Residual Dense Attention CNN & Too much complexity and requires more computational cost. & Less computational cost with 8.13M trainable parameters and 80 epochs of training. \\ \hline
        Yang et al. \cite{yang2022remote}  & AID, RSSCN7 & Node Entropy & No proper prediction task using deep learning approach. & Experimented with numerous DL architectures and configured the optimal model. \\ \hline
        Thirum-aladevi et al. \cite{thirumaladevi2023remote} & UCM,  SIRI-WHU & VGG, Support Vector Machine (SVM) & Low classification performance and consumed a very high number of iterations for training. & Has higher performance. Moreover, consumes only 80 epochs of training in total. \\ \hline
         Vinay-kumar et al. \cite{vinaykumar2023optimal}  & AID, NWPU, UCM & AlexNet, ResNet, BiLSTM, Whale Optimization  & No investigate alternative feature selection methods to confirm that the proposed method is the most effective. & Comparison with prior literature validates the effectiveness of our feature fusion method. \\ \hline
         Huang et al. \cite{huang2023semi} & NWPU-RESISC45, AID, RSSCN7, WHU-RS19 & Supervised alignment, Unsupervised alignment & Incorrect pseudo-labeled target samples due to performance bottleneck. & Solved performance bottleneck with soft voting. \\ \hline
         Jabbar et al. \cite{al2023ebola}& RSSCN7 & DarkNet-53, Ebola optimization & The paper lacks in-depth analysis and clarity regarding GCN implementation. & Provide a transparent and in-depth analysis of the methodology. \\ \hline
         Wang et al.\cite{wang2023large}&  
UCM, AID, NWPU-RESISC45 & LSCNet & Does not integrate transfer learning. Hence, requires 500 epochs to train. & We employ transfer learning, which reduces the training epoch to 80. \\ \hline
    \end{longtable}

An analysis of the research papers along with the limitations that our approach addresses is presented in Table \ref{tab:compare-lit}. The thorough analysis carried out in this study has exposed several significant shortcomings in the majority of the research works that have been done in the field. These research works specifically suffer from three major flaws, including a bottleneck caused by the integration of traditional feature extraction modules that are resistant to GPU acceleration, decreased accuracy, and excessive resource consumption during the training phase. While some research explored combining CNNs with vision transformers, they consume high resources to train due to the number of parameters of the architectures. Therefore, we developed a solution that alleviates the bottleneck and mitigates the high training costs of previous approaches. Our solution adopts a delicately designed architectural design that eliminates the bottleneck issue to alleviate these problems.  Our approach involves deploying four different models, each of which can be trained in parallel, and which together have a modest 8.1 million trainable parameters. The training regimen consists of 80 epochs, with each model receiving 20 epochs. Remarkably, this method leads to state-of-the-art performance, representing a significant advancement in the field. 

\section{Proposed Method}
\label{sec:method}
\begin{figure*}[h]
    \centering
    \includegraphics[width=\linewidth,height=4cm]{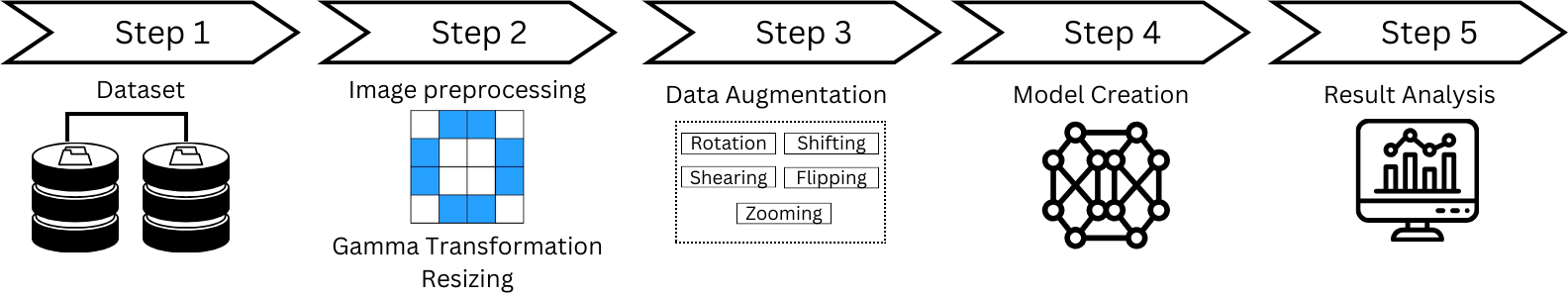}
    \caption{Flow diagram of the proposed method.}
    \label{fig:flow}
\end{figure*}

This section discusses the data preprocessing and augmentation steps as well as the proposed model architecture in detail. Figure \ref{fig:flow} presents an overview of the process.

\subsection{Data Preprocessing}
Image preprocessing is one of the crucial steps for enhancing image classification performance. As remote sensing images are captured from a distance, they often suffer from limited visibility, especially for small and dark objects. One of the common methods used to address this is gamma transformation \cite{sharma2022image}. In gamma transformation, as shown in Eq \ref{eq:gammatransform}, a non-linear function maps the input image $r$ to an output image $s$ by raising each pixel value to a power of $\gamma$ and multiplying it with a constant $c$. Since the images of both datasets are captured from the satellite, some images lack visibility. Therefore, when $\gamma$ is greater than 1, it intensifies the brightness of the images, leading to improved visualization of small, dark objects. Nevertheless, a higher $\gamma$ degrades the image quality by inflating the high-intensity pixels. Hence, after a meticulous experiment, the gamma transformation was performed with a $\gamma$ value of 1.1 and $c$ set to 1.
\begin{equation}
    \label{eq:gammatransform}
    s=c \times r^\gamma
\end{equation}

We have employed transfer learning to enhance the resource efficiency of the feature extraction process. These state-of-the-art models, trained on ImageNet, typically require images to be resized to 224$\times$224 for architectural compatibility \cite{magdy2023performance}. This standardized input size is used throughout our experiments. In addition, ViT adaptive patch embedding can tackle input sizes, whereas CNN models modify their adaptive pooling layers to ensure compatibility in the input dimension. However, since the original resolution of the images in both datasets is higher, downscaling to this size could lead to performance degradation due to the loss of fine details. To address this, we resized the images to 448$\times$448 using nearest-neighbor interpolation. This careful resizing ensured compatibility with the pretrained feature extractor by maintaining a consistent scale factor of two and preserving crucial details in the remote sensing images. Next, the pixel values were normalized by dividing each pixel by 255. Image normalization ensures faster convergence. Additionally, some real-time data augmentation was implemented to augment the training images. The augmentation methods include:

\begin{itemize}
    \item \textbf{Random Rotation:} Images were randomly rotated by 40 degrees clockwise or anticlockwise.
    \item \textbf{Random Horizontal and Vertical Shifts:} Images are arbitrarily shifted horizontally and vertically by a maximum of 20\%. 
    \item \textbf{Random Shearing:} Shear transformation applied to the input images with a maximum shear angle of 0.2.
    \item \textbf{Random Zooming:} Images were randomly zoomed in and out in a range of 20\%.
    \item \textbf{Horizontal Flipping:} Some randomly selected images were flipped horizontally. 
\end{itemize}

\subsection{Model Architecture}
\begin{figure*}[h]
    \centering
    \includegraphics[height=9cm, width=\linewidth]{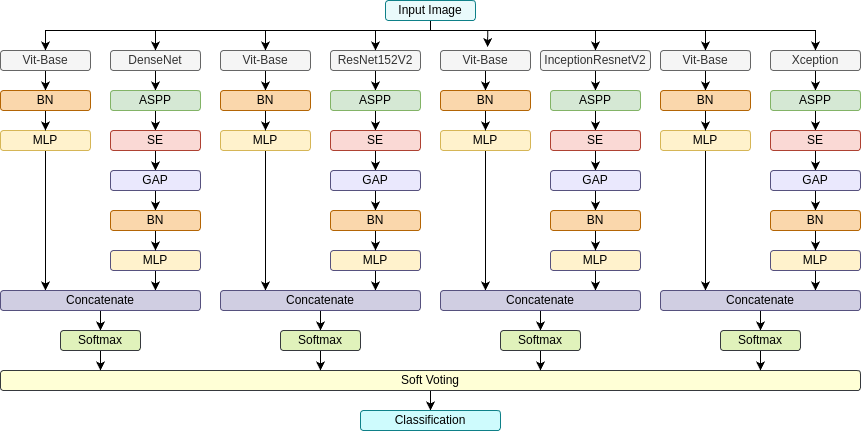}
    \caption{Proposed model architecture.}
    \label{fig:model}
\end{figure*}

As shown in Figure \ref{fig:model}, the proposed model ensembles four fusion models through a soft voting mechanism. Each fusion model consists of two streams. The first stream is the transformer stream, and the second stream is the CNN stream. The subsequent sections describe the architecture of the streams along with the fusion process. 

\subsubsection{Transformer Stream}
The transformer stream incorporated a pretrained ViT-Base model trained on the ImageNet1K dataset. ViT-Base is the smallest ViT architecture proposed by Dosovitskiy et al. \cite{dosovitskiy2020image}. The pretrained transformer feature extractor then followed a batch normalization (BN) to effectively reduce the training process. The BN layer has a momentum of 0.99 and an epsilon of 0.001 to enhance the training stability and generalization capability. Right after the BN, a Multi-Layer Perceptron (MLP) was added that consisted of three fully connected (FC) layers of 512, 256, and 121 neurons respectively with ReLU activation function. The first FC layer applied L2 kernel regularization with a coefficient of 0.016, L2 activity regularization with a coefficient of 0.006, and L1 bias regularization with a coefficient of 0.006. The gradual reduction in the number of neurons across the MLP layers was designed to progressively distill and refine the learned features.
\begin{figure}[h]
    \centering
    \includegraphics[width=\linewidth,height=8cm]{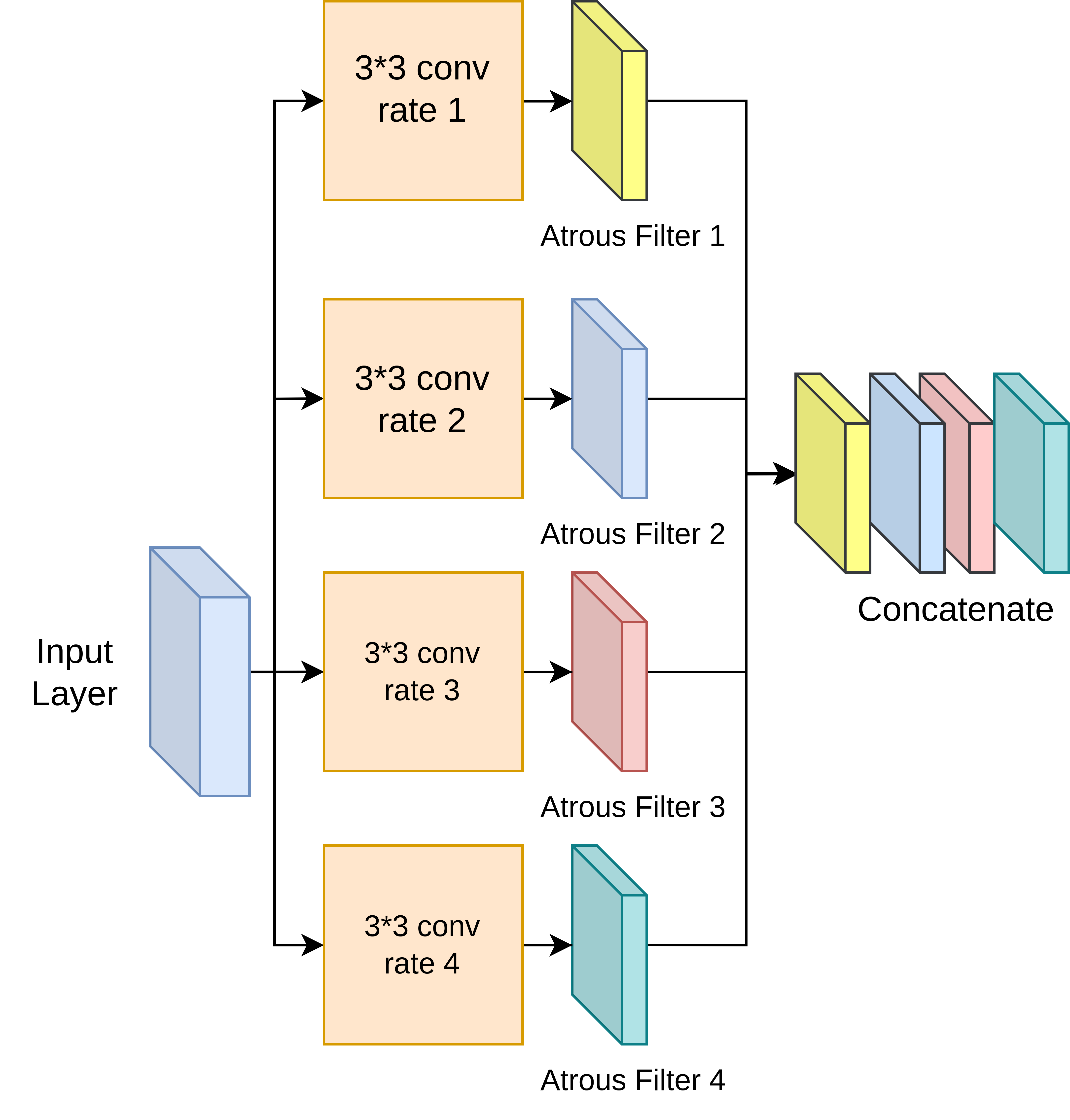}
    \caption{The ASPP module used in the model.}
    \label{fig:ASPP}
\end{figure}

\subsubsection{CNN Stream}
Each fusion model-specific CNN stream included a pretrained CNN feature extractor for extracting distinguishing features from the input images. After the CNN feature extractor, an Atrous Spatial Pyramid Pooling (ASPP) module was integrated for extracting multi-scale features. ASPP captures multi-scale contextual information by applying atrous (dilated) convolutions with different dilation rates. The change in the dilation rates expands the receptive field without increasing the kernel size, which ensures resource efficiency. Since remote sensing images have both fine and large features that contribute to the classification process, ASPP is essential for accurately detecting and analyzing these features. As presented in Figure \ref{fig:ASPP}, the ASPP module included several parallel dilated convolutions ranging in dilation rates (2, 3, 5, and 7) on a 3$\times$3 kernel. According to the experiment, the ASPP module dramatically increased the classification performance. Additionally, since the feature extractor was trained on ImageNet, which contains imagery dissimilar to the remote sensing data, some feature maps do not contribute equally to the classification process. Therefore a squeeze-and-excitation (SE) block was incorporated after the ASPP module to emphasize the important feature maps and reduce the influence of those with lesser contribution. SE block is an attention mechanism that imposes channel attention on the feature maps \cite{hu2018squeeze}. 

The attention block operates through three major steps: compression, excitation, and recalibration. During the compression step, depicted in Eq \ref{eq:compression}, the spatial dimensions of a feature map are reduced using global average pooling. Let $F_c$ represent a feature map with height $H$ and width $W$. The channel summary, denoted by scalar $c_c$, is computed by averaging the values of the channel $F_c$.

\begin{equation}
\label{eq:compression}
    c_c = \frac{1}{H \times W}\sum_{i=1}^{H}\sum_{j=1}^{W}F_c(i,j)
\end{equation}

In the excitation step, shown in Eq \ref{eq:excitation}, the scalar $c_c$ is processed through two fully connected layers. Let $W_1$ and $W_2$ denote the weights of the first and second layers, respectively. The first layer uses ReLU activation, while the second layer uses a sigmoid activation function. The excitation phase produces the channel weight $e_c$, which is then used to scale the feature map, resulting in the recalibrated feature map $R_i$. The recalibration process is defined in Eq. \ref{eq:recalibration}.

\begin{equation}
\label{eq:excitation}
    e_c = \sigma(W_2 \times ReLU(W_1 \times c_c))
\end{equation}

\begin{equation}
\label{eq:recalibration}
    R_i = e_c \times F_c
\end{equation}

For converting 2D feature maps into 1D feature vectors, global average pooling (GAP) was employed. GAP returns a single value by taking the average of the feature map. As per the experiment, GAP layer increases the classification accuracy compared to the flattening layer by remarkably reducing the number of parameters. The feature vectors are then passed through a BN layer and an MLP block. The configuration of the BN and MLP is the same as that in the transformer stream. 

\subsubsection{Fusion Process} 
This fusion process blends the complementary strengths of both streams, taking advantage of the spatial information captured by the CNN stream and the global context modeling provided by the transformer stream. In this process, the outputs from the transformer stream and the CNN stream were concatenated and passed through softmax layers for classification. Therefore, each fusion model had the classification ability. The four fusion models vary in only the CNN feature extractor module. Four pretrained CNN models were integrated in fusion models namely DenseNet121 \cite{huang2017densely}, ResNet152V2 \cite{he2016deep}, InceptionResnetV2 \cite{szegedy2017inception} and Xception \cite{chollet2017xception}. 

\subsubsection{Soft Voting and Classification}
Soft voting is an ensemble technique used to combine predictions from various models \cite{kim2022inspection}. Instead of making discrete predictions, each model in soft voting provides class probabilities. To obtain a final prediction, these probabilities are then averaged, which typically produces more reliable and accurate classification decisions. The proposed model performed soft voting on the four fusion models that were constructed previously. Since in our scenario, summing up the class probabilities and calculating the average probability makes no difference in the overall classification performance, we take the summation of the four predictions and the class label with the highest probability is declared as a prediction. 

\subsubsection{Hyperparameters}
Each fusion model was compiled on Adam optimizer and the categorical cross-entropy loss function. The training process of all the models lasted for 20 epochs. In total, the four models consumed 80 epochs of training. The learning rate for each model was set to 0.001. The batch size for all the models was 64. The weights and biases of the feature extractor of the fusion models were initialized with ImageNet pre-trained weights and set to non-trainable.

\section{Results and Discussion}
\label{sec:results}
This section discusses the results obtained from the proposed model. In addition to that, it also presents an ablation study.

\subsection{Dataset Description}

\begin{figure*}
    \centering
    \includegraphics[width=\textwidth,height=7cm]{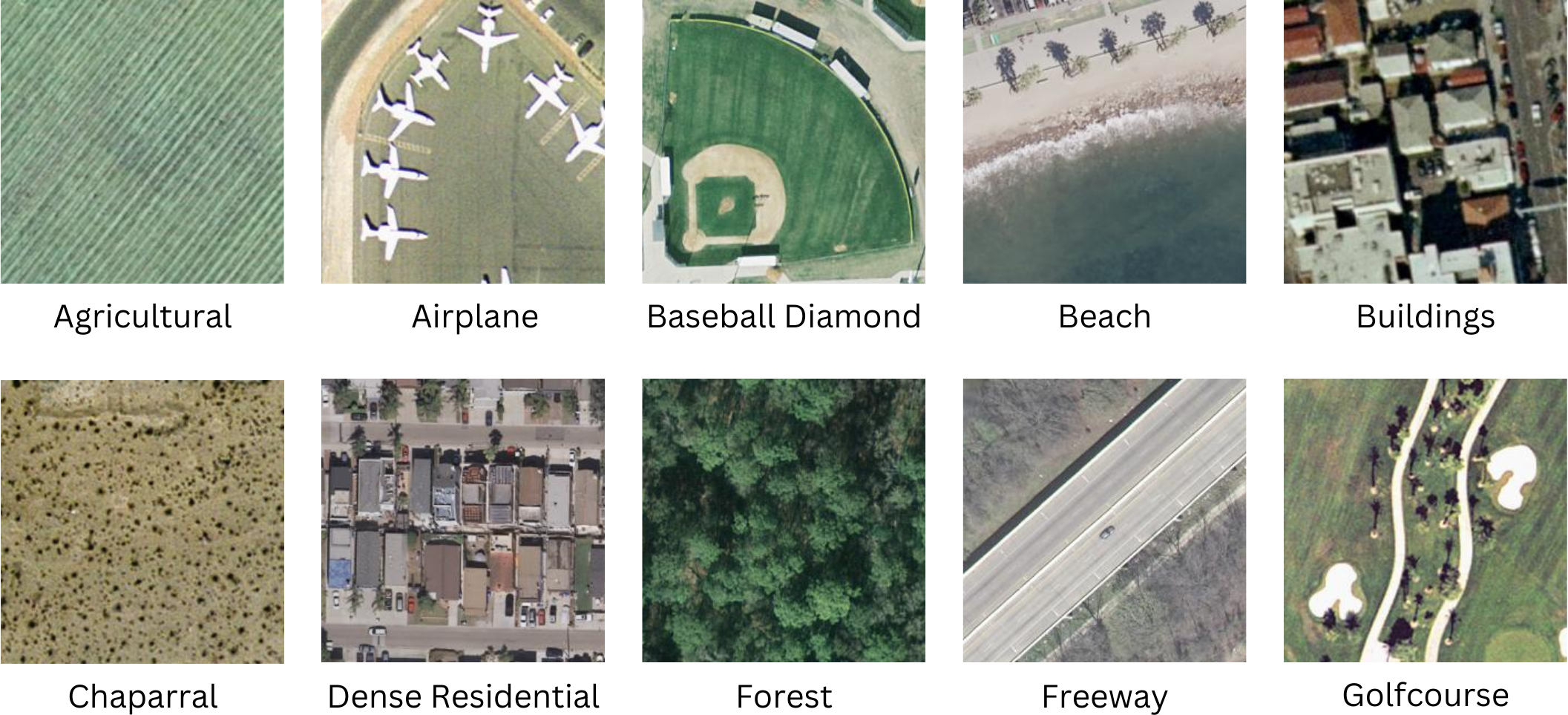}
    \caption{Samples from the UCM dataset.}
    \label{fig:UCM}
\end{figure*}

\begin{figure*}
    \centering
    \includegraphics[width=\textwidth,height=7cm]{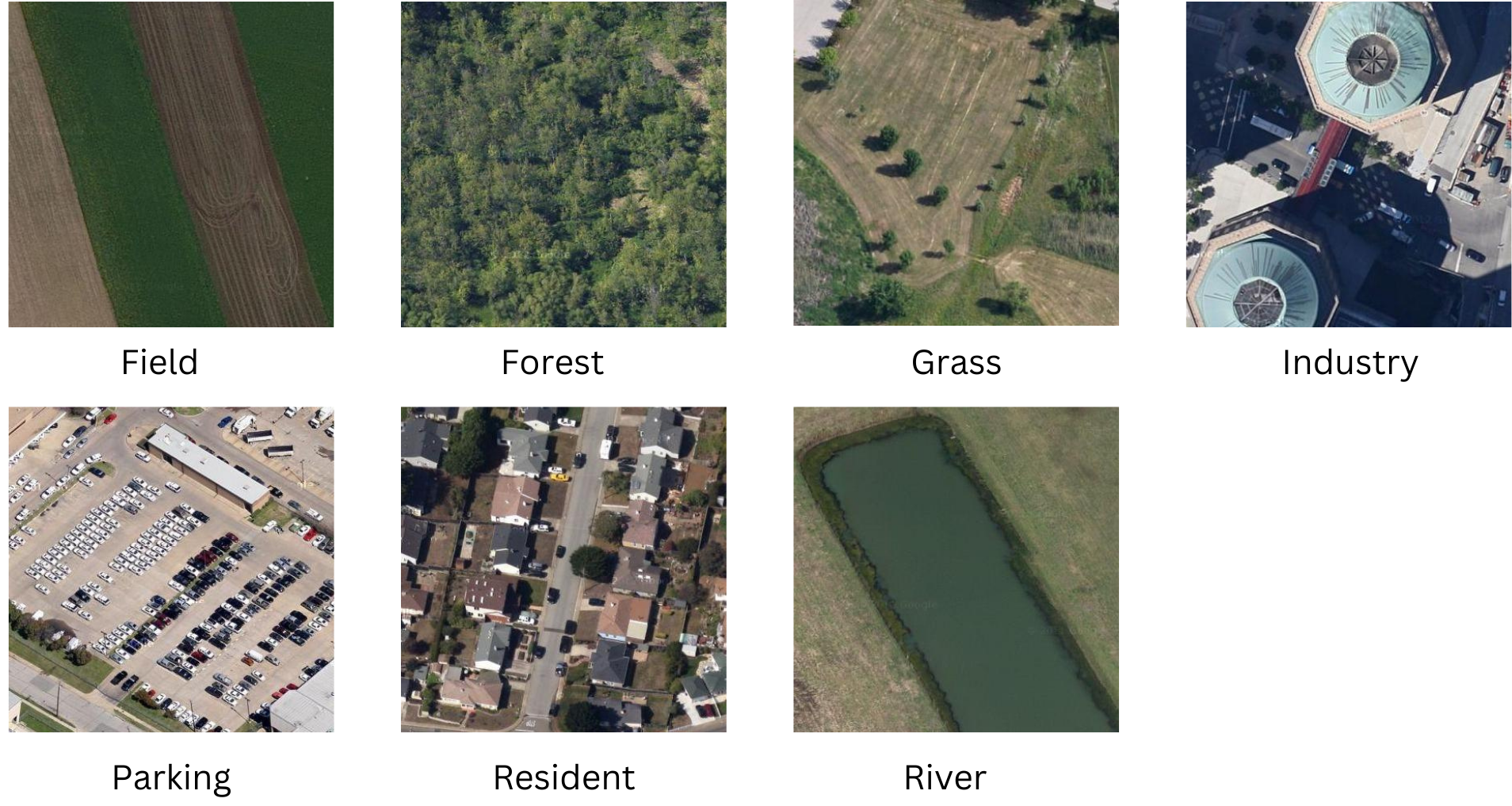}
    \caption{Samples from the RSSCN7 dataset.}
    \label{fig:RSSCN7}
\end{figure*}

\begin{figure*}
    \centering
    \includegraphics[width=\textwidth,height=7cm]{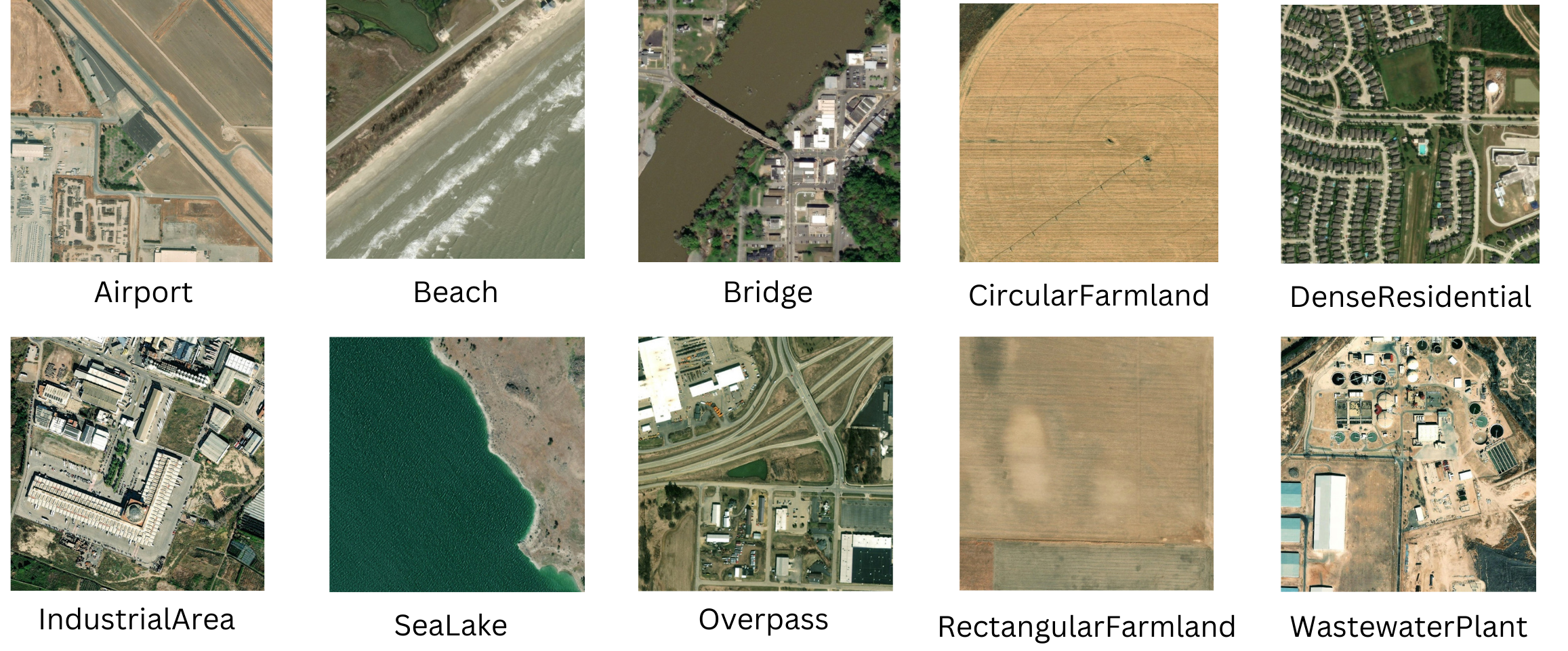}
    \caption{Samples from the MSRSI dataset.}
    \label{fig:MSRSI}
\end{figure*}

For evaluating the model's performance, three benchmark datasets are selected. The first dataset is the UC Merced Land Use (UCM) Dataset, one of the most popular datasets in the field of remote sensing, consisting of 21 classes of various land cover types formed by the US Geological Survey \cite{yang2010bag}. With a total of 2100 images of with 100 images from each class, this fully balanced dataset contains high-resolution aerial images captured from a satellite of size 256 $\times$256 pixels. RSSCN7 dataset, on the other hand, is comprised of 2800 images of 7 classes acquired from Google Earth throughout multiple years time span \cite{ali2018hybrid}. The images of this dataset are of size 400 $\times$400. Lastly, the MSRSI dataset consists of five types of remote sensing images, including VHR, MSI, SAR, MAP, and PAN, collected from sources like Google Maps \cite{sun2023consistency}. For this experiment, we considered only the VHR type of images to match the other datasets. The dataset contains 30,000 VHR images with 15 class labels, evenly distributed across each class. A sample of the UCM, RSSCN7, and MSRSI datasets is presented in Figure \ref{fig:UCM}, \ref{fig:RSSCN7}, and \ref{fig:MSRSI}, respectively.

\subsection{Result Analysis}
Before the final soft voting, each fusion model was trained and evaluated on the same datasets. The accuracy of the models paired with different transformers is presented in Table \ref{tab:individual}. Each model was trained using the same hyperparameters. The results show that the CNNs paired with ViT (ViT Base) outperformed the Swin Transformer (SwinT) \cite{liu2021swin} and the Data-efficient Image Transformer (DeiT) \cite{touvron2021training}. Additionally, the fusion model constructed with ViT Base and DenseNet121 performed the best on the UCM dataset, while pairing ResNet152V2 with ViT achieved the highest accuracy on the RSSCN7 and MSRSI datasets. However, combining Xception with transformers resulted in lower mean and median accuracy compared to DenseNet121 and ResNet152V2. This difference in performance could be attributed to the fact that both DenseNet121 and ResNet152V2 propagate previously extracted features through multiple layers to retain minute details important for classification. Xception, while propagating features, also includes a pointwise convolution block to shrink the size of feature maps and reduce computational cost. This process may inadvertently discard fine features essential for the decision-making process in remote sensing images.

\begin{table*}[t]
\caption{Performance of each fusion model before soft voting with different transformer models. RSS refers to the RSSCN7 dataset and MSR refers to the MSRSI dataset.}
\centering
\small
\begin{adjustbox}{max width=\textwidth}
\begin{tabular}{|p{2cm}|p{1cm}|p{1cm}|p{1cm}|p{1cm}|p{1cm}|p{1cm}|p{1cm}|p{1cm}|p{1cm}|p{2.2cm}|p{1cm}|}
\hline
\multirow{2}{*}{\textbf{Model}} &
\multicolumn{3}{c|}{\textbf{ViT}} &
\multicolumn{3}{c|}{\textbf{SwinT}} &
\multicolumn{3}{c|}{\textbf{DeiT}} &
\multirow{2}{*}{\shortstack{\textbf{Trainable}\\\textbf{Parameters}}} &
\multirow{2}{*}{\textbf{Epoch}} \\
\cline{2-10}
& \textbf{UCM} & \textbf{RSS} & \textbf{MSR}
& \textbf{UCM} & \textbf{RSS} & \textbf{MSR}
& \textbf{UCM} & \textbf{RSS} & \textbf{MSR}
& & \\
\hline
DenseNet-121 & 95.47 & 91.07 & 91.03 & 95.00 & 90.54 & 82.17 & 94.05 & 89.82 & 77.92 & 1.5M & 20 \\
\hline
ResNet-152V2 & 92.38 & 91.25 & 91.95 & 91.42 & 91.61 & 84.77 & 93.33 & 92.50 & 85.50 & 1.9M & 20 \\
\hline
Inception ResNetV2 & 92.86 & 89.46 & 92.25 & 91.66 & 80.00 & 83.58 & 90.71 & 87.86 & 84.17 & 2.3M & 20 \\
\hline
Xception & 91.90 & 87.32 & 89.75 & 88.81 & 89.29 & 87.15 & 89.82 & 91.19 & 82.03 & 2.3M & 20 \\
\hline
\end{tabular}
\end{adjustbox}
\label{tab:individual}
\end{table*}

\begin{figure*}
    \centering
    \includegraphics[width=\textwidth,height=11cm]{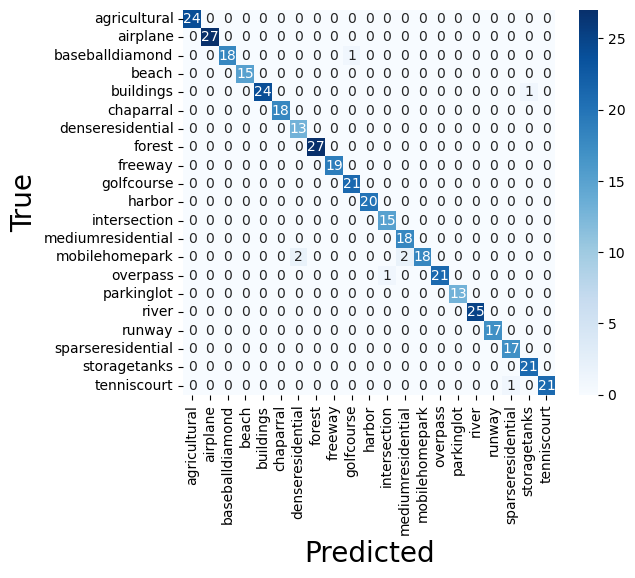}
    \caption{Confusion matrix of the proposed model on the UCM dataset.}
    \label{fig:ucm-cm}
\end{figure*}

\begin{figure*}
    \centering
    \includegraphics[width=\textwidth,height=9cm]{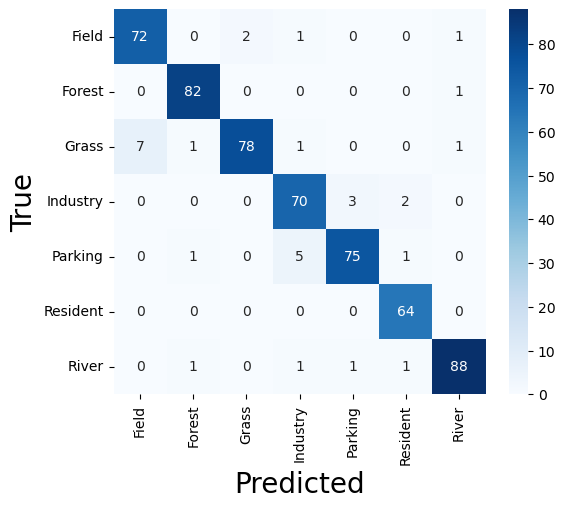}
    \caption{Confusion matrix of the proposed model on the RSSCN7 dataset.}
    \label{fig:rsscn7-cm}
\end{figure*}

\begin{figure*}
    \centering
    \includegraphics[width=\textwidth,height=11cm]{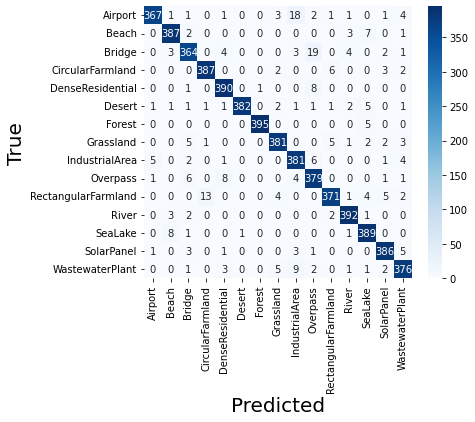}
    \caption{Confusion matrix of the proposed model on the MSRSI dataset.}
    \label{fig:msrsi-cm}
\end{figure*}

Various evaluation metrics were considered when evaluating the final model. The confusion matrix presents a detailed overview of the model's performance at a granular level. The confusion matrices for the three datasets on the proposed model are presented in Figure \ref{fig:ucm-cm}, \ref{fig:rsscn7-cm} and \ref{fig:msrsi-cm}. The metrics illustrate that the model can classify the majority of the images correctly. On the UCM dataset, two mobile home parks were misclassified as dense residential areas, and two others were misclassified as medium residential parks. This misclassification is primarily attributed to high inter-class similarity. Additionally, no other classes in the UCM dataset experienced more than one misclassification. On the RSSCN7 dataset, however, seven grass images were misclassified as fields. This result indicates that global features negatively impacted the classification of some grass images, as both grass and field images share similarities in their global context. Since four models are trained individually and their convergence shows no signs of overfitting, the absence of cross-validation can be mitigated by the observed stability and consistency. On the MSRSI dataset, the highest misclassification was reported in labeling bridges as overpasses. Additionally, some airport images were misclassified as industrial areas, and rectangular farmland as circular farmland. These results show that the model struggles in instances with high inter-class similarity.

\begin{figure*}[h!]
    \centering
    \begin{subfigure}[b]{0.32\textwidth}
        \centering
        \includegraphics[height=3.8cm]{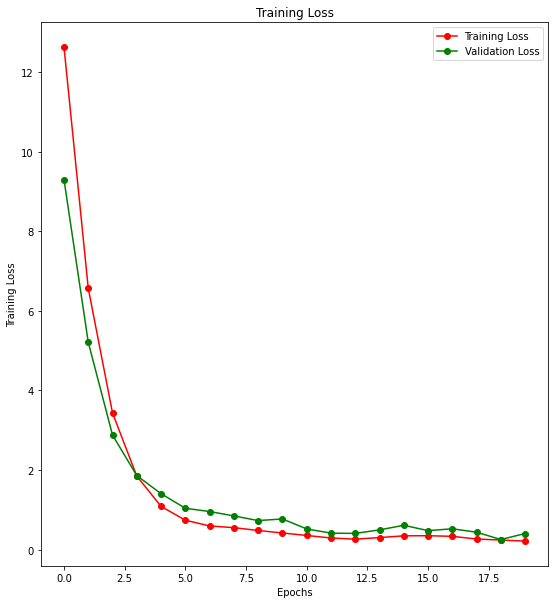}
        \caption{UCM: ViT Base + DenseNet121}
        \label{fig:ucm_densenet121}
    \end{subfigure}
    \hfill
    \begin{subfigure}[b]{0.32\textwidth}
        \centering
        \includegraphics[height=3.8cm]{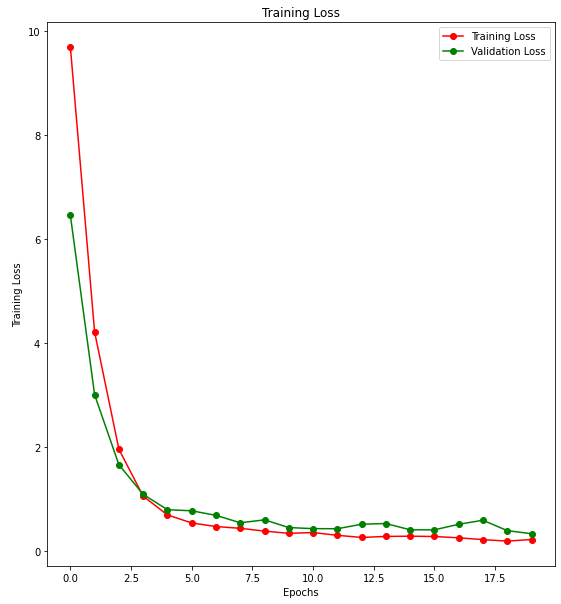}
        \caption{RSSCN7: ViT Base + DenseNet121}
        \label{fig:rsscn_densenet}
    \end{subfigure}
    \hfill
    \begin{subfigure}[b]{0.32\textwidth}
        \centering
        \includegraphics[height=3.8cm]{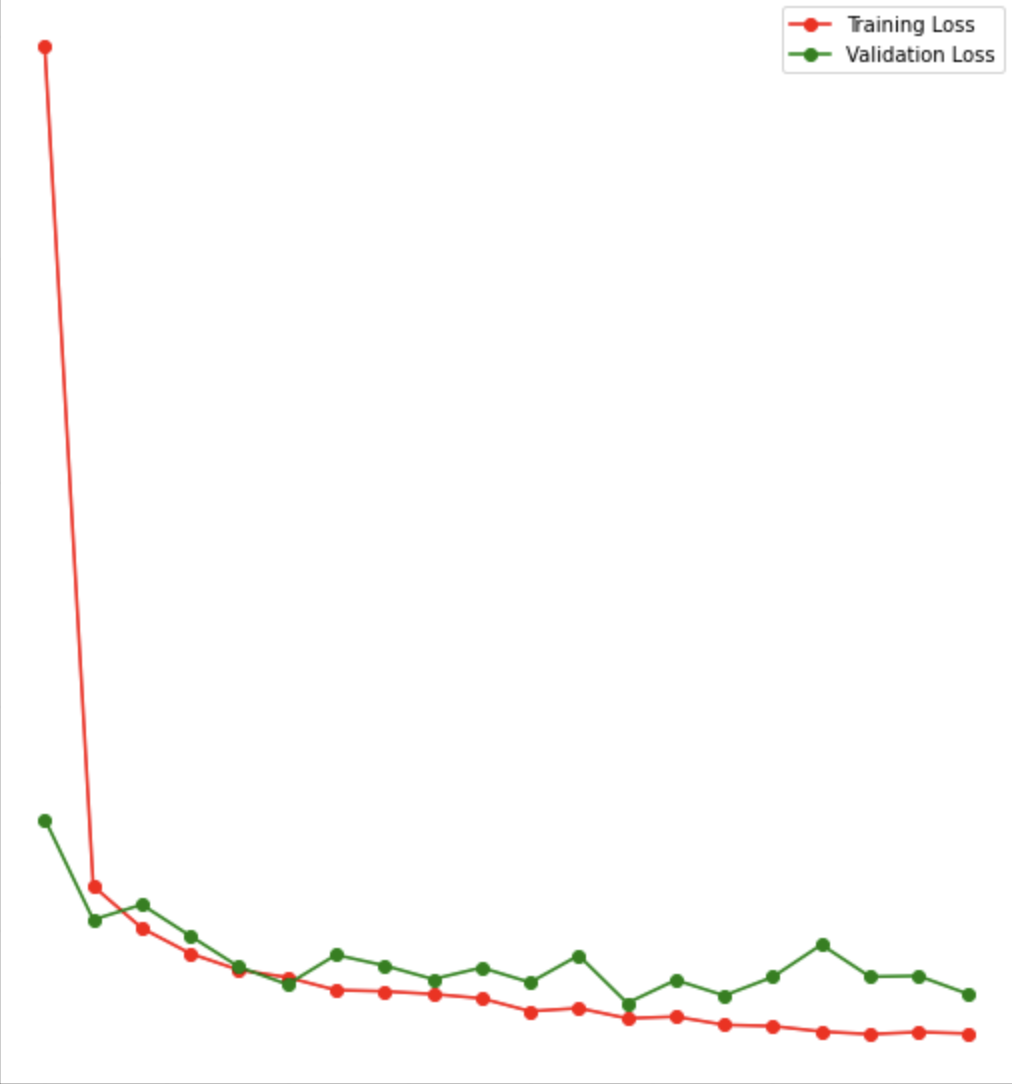}
        \caption{MSRSI: ViT Base + DenseNet121}
        \label{fig:msris_densenet}
    \end{subfigure}
    
    \vskip\baselineskip
    
    \begin{subfigure}[b]{0.32\textwidth}
        \centering
        \includegraphics[height=3.8cm]{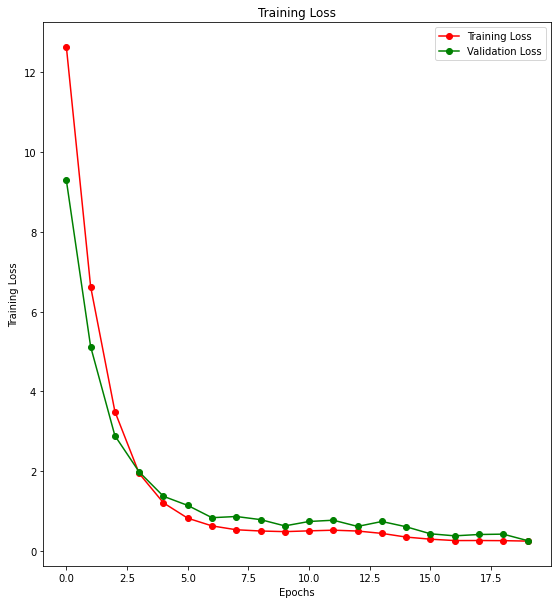}
        \caption{UCM: ViT Base + InceptionResNetV2}
        \label{fig:ucm_iresnet}
    \end{subfigure}
    \hfill
    \begin{subfigure}[b]{0.32\textwidth}
        \centering
        \includegraphics[height=3.8cm]{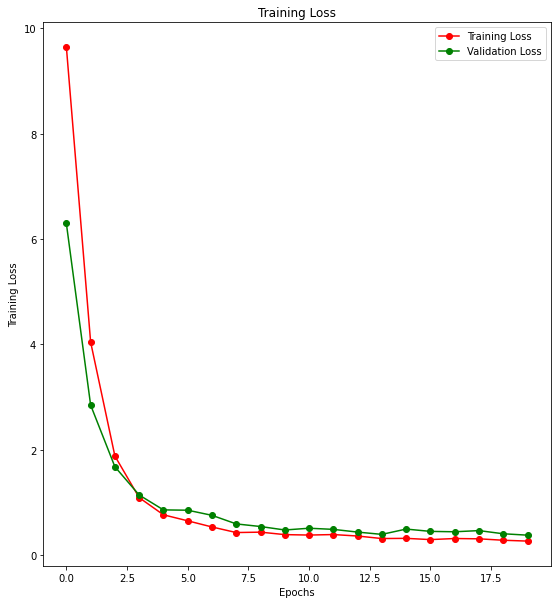}
        \caption{RSSCN7: ViT Base + InceptionResNetV2}
        \label{fig:rsscn_iresnet}
    \end{subfigure}
    \hfill
    \begin{subfigure}[b]{0.32\textwidth}
        \centering
        \includegraphics[height=3.8cm]{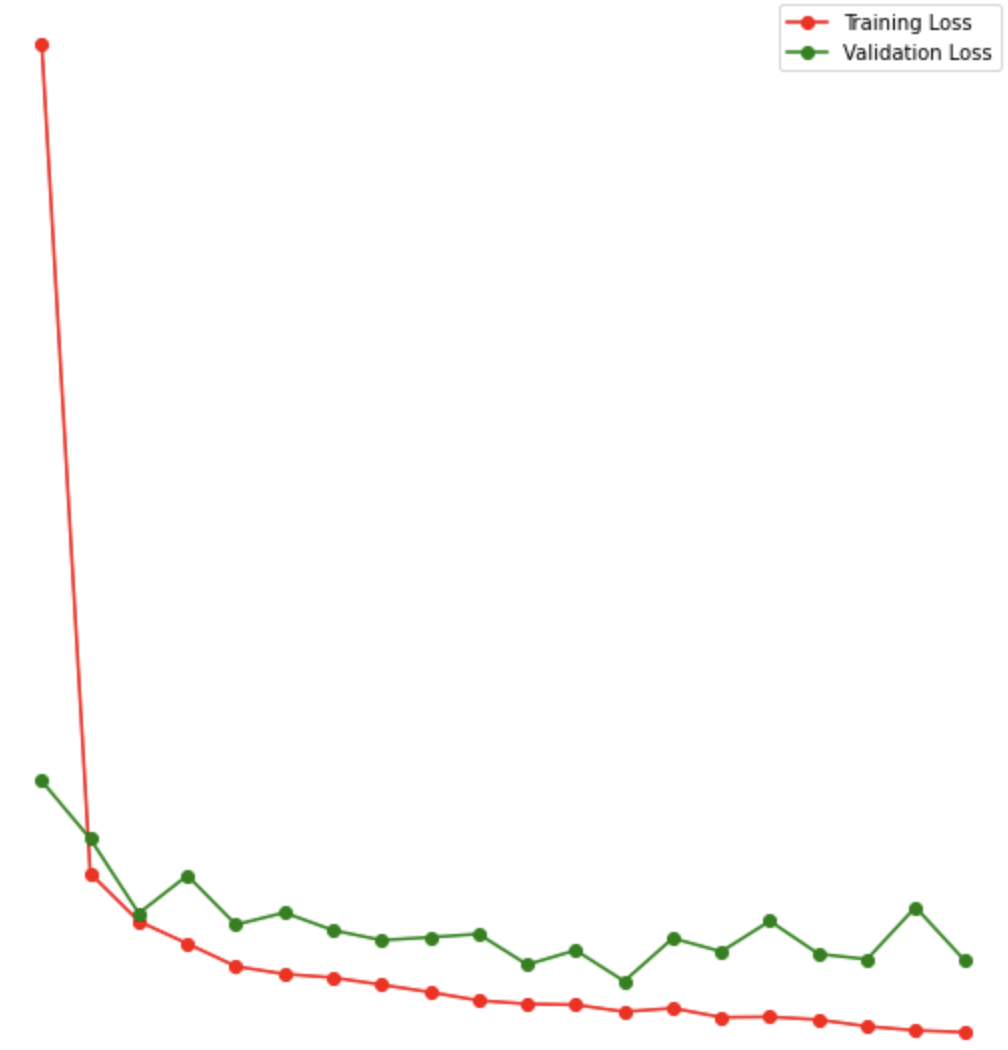}
        \caption{MSRSI: ViT Base + InceptionResNetV2}
        \label{fig:msris_iresnet}
    \end{subfigure}
    
    \vskip\baselineskip
    
    \begin{subfigure}[b]{0.32\textwidth}
        \centering
        \includegraphics[height=3.8cm]{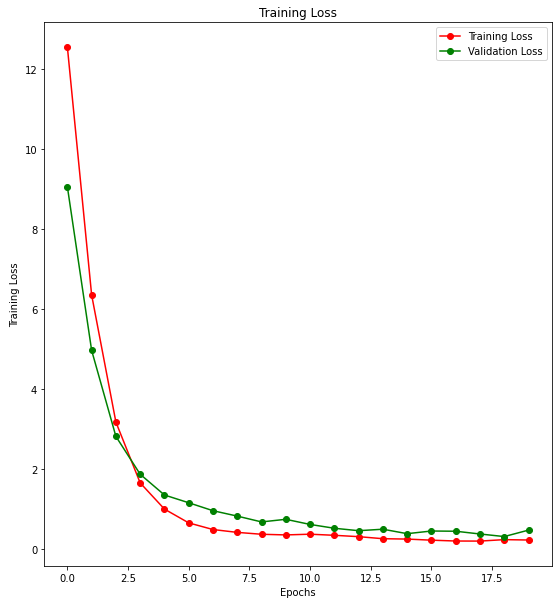}
        \caption{UCM: ViT Base + ResNet152v2}
        \label{fig:ucm_resnet152v2}
    \end{subfigure}
    \hfill
    \begin{subfigure}[b]{0.32\textwidth}
        \centering
        \includegraphics[height=3.8cm]{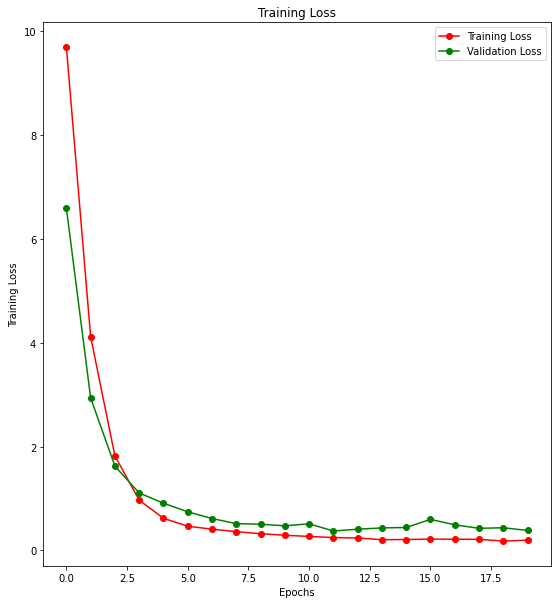}
        \caption{RSSCN7: ViT Base + ResNet152v2}
        \label{fig:rsscn_resnet152v2}
    \end{subfigure}
    \hfill
    \begin{subfigure}[b]{0.32\textwidth}
        \centering
        \includegraphics[height=3.8cm]{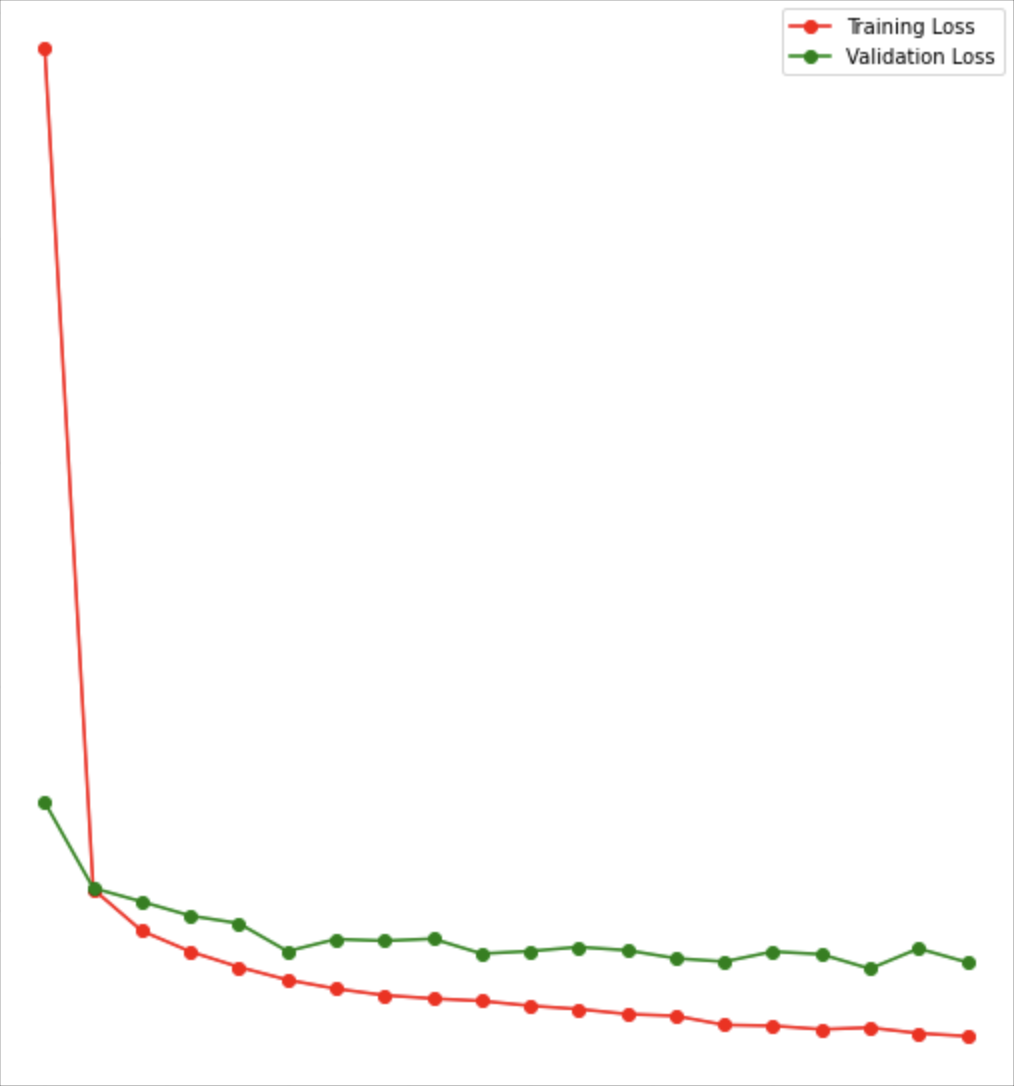}
        \caption{MSRSI: ViT Base + ResNet152v2}
        \label{fig:msris_resnet152v2}
    \end{subfigure}
    
    \vskip\baselineskip
    
    \begin{subfigure}[b]{0.32\textwidth}
        \centering
        \includegraphics[height=3.8cm]{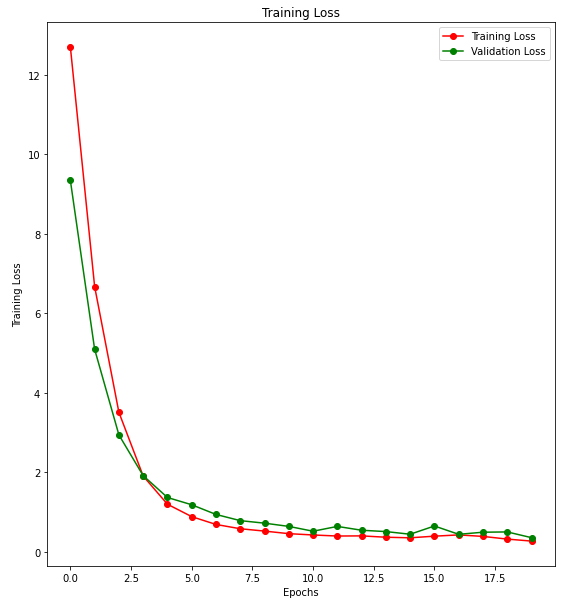}
        \caption{UCM: ViT Base + Xception}
        \label{fig:ucm_xception}
    \end{subfigure}
    \hfill
    \begin{subfigure}[b]{0.32\textwidth}
        \centering
        \includegraphics[height=3.8cm]{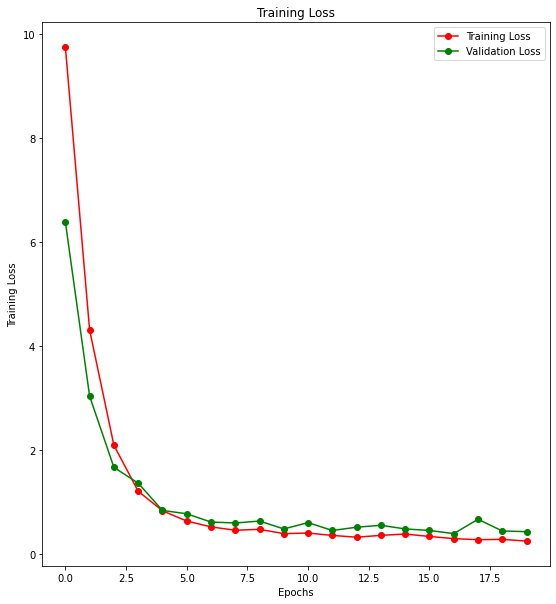}
        \caption{RSSCN7: ViT Base + Xception}
        \label{fig:rsscn_xception}
    \end{subfigure}
    \hfill
    \begin{subfigure}[b]{0.32\textwidth}
        \centering
        \includegraphics[height=3.8cm]{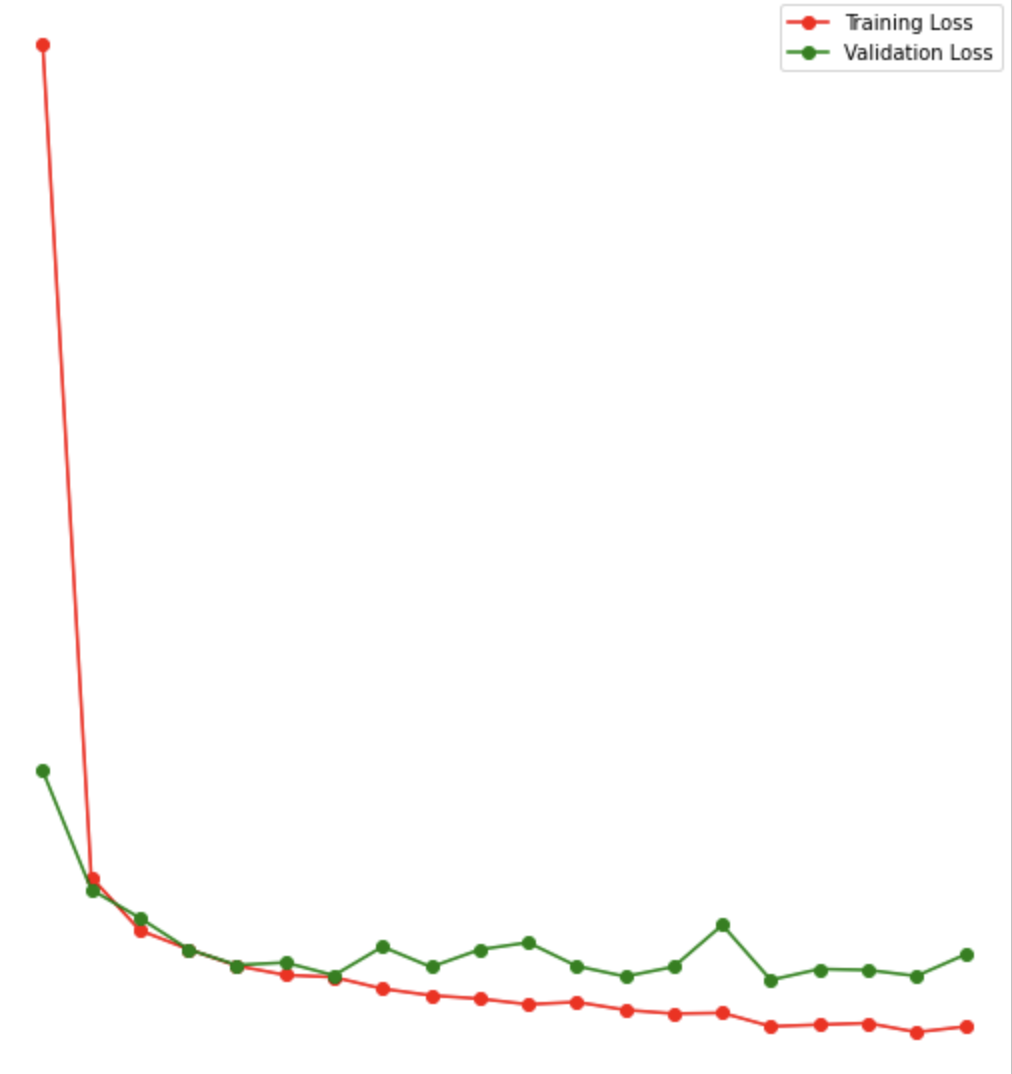}
        \caption{MSRSI: ViT Base + Xception}
        \label{fig:msris_xception}
    \end{subfigure}

    \caption{Loss graphs of four fusion models used for ensembling on the UCM, RSSCN7, and MSRSI datasets.}
    \label{fig:all_loss_graphs}
\end{figure*}

Figure \ref{fig:all_loss_graphs} presents the loss graphs (cross-entropy loss) on the UCM, RSSCN7, and MSRSI datasets. The loss graphs indicate overall smooth convergence as the training progresses, showing stable learning behavior across the models. This pattern indicates a reduced likelihood of overfitting during the training process.

To evaluate the efficacy of the proposed method, we also employed various performance metrics, such as precision, recall, F1-score, false positive rate (FPR), true positive rate (TPR), and Matthews Correlation Coefficient (MCC) \cite{vujovic2021classification}. The performance assessment of our proposed method for both datasets is shown in Table \ref{tab:performance}. The table illustrates that the proposed method achieved an overall adequate performance in each metric, especially for the UCM dataset. It consistently accomplishes a remarkable true positive rate of 100\% and a false positive rate of 0\%. Additionally, the method demonstrated a high accuracy of 98.10\% for the UCM dataset, while still performing well on the RSSCN7 dataset with an accuracy of 94.46\%. MCC, regarded as the most reliable metric among the seven compared \cite{chicco2021matthews}, further supports these results, indicating robust classification performance for both datasets, with values of 98.00\% and 93.55\% for the UCM and RSSCN7 datasets, respectively. Similarly, the method achieves an accuracy of 95.45\% on the MSRSI dataset and an MCC of 95.13\%. These results underscore the model's effectiveness, where it excelled across all evaluated metrics. The high performance on all three datasets suggests the generalizability of the proposed method.

\begin{figure*}
    \centering
    \includegraphics[width=\textwidth,height=11cm]{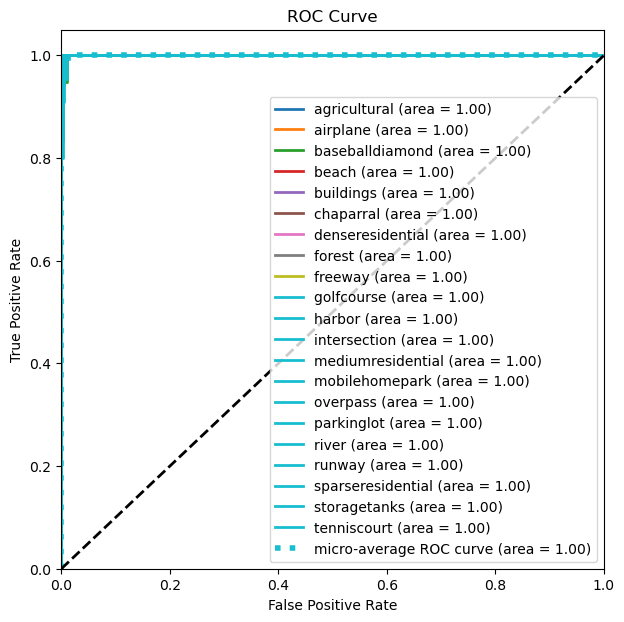}
    \caption{ROC curve of the proposed model on the UCM dataset.}
    \label{fig:ucm-roc}
\end{figure*}

\begin{figure*}
    \centering
    \includegraphics[width=\textwidth,height=9cm]{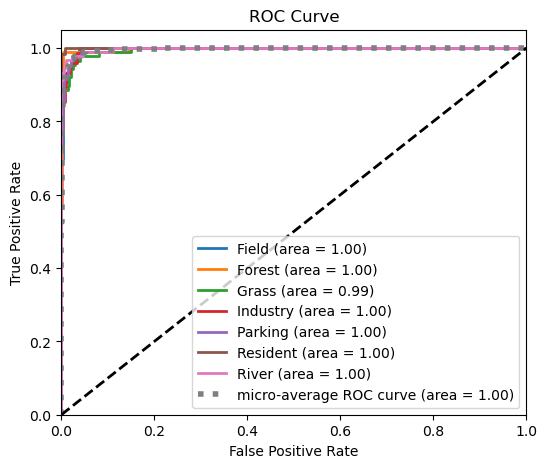}
    \caption{ROC curve of the proposed model on the RSSCN7 dataset.}
    \label{fig:rsscn7-roc}
\end{figure*}

\begin{figure*}
    \centering
    \includegraphics[width=\textwidth,height=10cm]{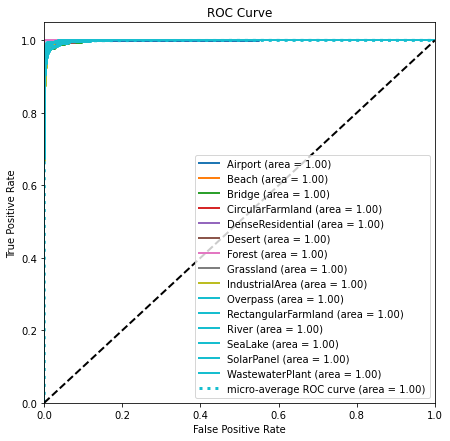}
    \caption{ROC curve of the proposed model on the MSRSI dataset.}
    \label{fig:msrsi-roc}
\end{figure*}

\begin{table}[h]
    \centering
    \caption{Performance analysis of the proposed approach.}
    \begin{tabular}{|p{.30\linewidth}|p{.15\linewidth}|p{.15\linewidth}|p{.15\linewidth}|}
    \hline
        \textbf{Measure} & \textbf{UCM} & \textbf{RSSCN7} & \textbf{MSRSI}\\ \hline
        Accuracy & 98.10 & 94.46 & 95.45\\ \hline
        Precision & 98.31 & 94.61 & 95.49\\ \hline
        Recall & 98.10 & 94.46 & 95.45\\ \hline
        F1-Score & 98.11 & 94.48 & 95.45  \\ \hline
        True Positive Rate & 100.00 & 92.13 & 97.97 \\ \hline
        False Positive Rate & 0.00 & 4.00 & 4.18 \\ \hline
        MCC & 98.00 & 93.55& 95.13 \\ \hline
    \end{tabular}
    \label{tab:performance}
\end{table}

Another very effective evaluation parameter is the Receiver Operating Characteristic (ROC) curve. It presents the model's ability to differentiate between true positive and false positive instances. The ROC curve for the proposed model is present in Figure \ref{fig:ucm-roc}, \ref{fig:rsscn7-roc}, and \ref{fig:msrsi-roc}. The micro average ROC curve on all datasets is 1.0 which is the highest value. This proves the model's strong discriminate ability. Moreover, in terms of computation, the proposed solution utilizes transfer learning that makes the models train significantly fewer parameters (8.13M) and consumes fewer training epochs (80 epochs in total). 

\begin{table}[h]
    \centering
    \caption{Comparison with the state-of-the-art methods.}
    \begin{tabular}{|p{2cm}|p{1.2cm}|p{1.7cm}|p{1.7cm}|p{2cm}|p{2.2cm}|p{1.7cm}|}
    \hline
         \textbf{Model Name} & \textbf{UCM} & \textbf{RSSCN7} & \textbf{MSRSI} & \textbf{Total parameters} & \textbf{Trainable Parameters} & \textbf{Training Epoch} \\ \hline
         Xception & 94.05 & 87.32 & 82.13 & 22.62 M & 0.57 M & 100 \\ \hline
        Inception-ResnetV2 & 92.86 & 88.57& 80.71 & 54.9 M & 0.56 M & 100 \\ \hline
        XCiT & 80.95 & 80.89 & 71.81 & 84 M & 0.56 M & 100 \\ \hline
         Swin Transformer & 95.95 & 91.79& 88.10 & 195.95 M & 0.95 M & 100 \\ \hline
        ViT Base & 95.71 & 89.11 & 86.55 & 86.36 M & 0.95 M & 100 \\ \hline
        DeiT & 90.48 & 91.07 & 84.15 & 86.65 M & 0.56 M & 100 \\ \hline
         CLIP - ResNet50 & 46.43 & 49.82 & 34.46 & 244 M & 0 & 0 \\ \hline
      CLIP - ViT-L/14 & 71.42 & 58.75 & 68.48 & 891 M  & 0 & 0 \\ \hline
      SigLIP  & 72.14 & 59.64 & 62.21 & 400 M  & 0 & 0 \\ \hline
      P$^{2}$FEViT & 80.48 & 84.11& 59.37 & 5.63 M & 1.58 M & 100 \\  \hline
        Proposed & 98.10 & 94.46 & 95.45  & 495.5 M & 8.1 M & 80 \\ \hline
    \end{tabular}
    \label{tab:model_comp}
\end{table}

 Although making a direct comparison among various research articles is challenging due to the absence of detailed resource consumption information, we have compared our results with state-of-the-art image classifiers. CNNs such as Xception and InceptionResNetV2, fine-tuned in a similar fashion to the CNN stream of our proposed classifier, have achieved 94.05\% and 92.86\% accuracy on the UCM dataset, 87.32\% and 88.57\% accuracy on the RSSCN7 dataset, and 82.13\% and 88.57\% on the MSRSI dataset, respectively. Transformer models such as Swin-T and ViT-Base outperformed other variants like XCiT \cite{ali2021xcit} and DeiT by a noticeable margin. Additionally, we have compared our model with Contrastive Language-Image Pretraining (CLIP) \cite{radford2021learning}, a powerful neural network that has outperformed existing CNN and transformer models on various image recognition tasks. The model learns to associate textual descriptions with images by optimizing two encoders (a text and an image encoder). It maximizes the similarity between an image and its corresponding text while minimizing the similarity to unrelated pairs. Once trained, CLIP can generalize to new tasks without additional training, which enables the model to perform zero-shot image classification. We evaluated the model by changing the image encoder, but it failed to produce acceptable accuracy. Notably, although the model does not require any fine-tuning, which is reflected in the trainable parameters, it has one of the highest numbers of parameters due to the integration of two heavy encoder models. Similar to CLIP, we have evaluated another newly developed zero-shot image classifier SigLIP \cite{zhai2023sigmoid}. Similar to CLIP, it fails to achieve notable performance. Lastly, we have evaluated  P$^{2}$FEViT, a hybrid transformer-CNN architecture developed by Wang et al. \cite{wang2023p} for remote sensing image classification. Table \ref{tab:model_comp} presents a direct comparison with the models discussed. All the models (except CLIP and SigLIP) were trained for a total of 100 epochs using the same GPU, the P100. Despite the extensive training epochs, these models underperformed compared to our soft voting fusion model, which achieved superior performance after only 20 epochs of training. Although the proposed model has over 495 million parameters, the number of trainable parameters is only 8.1 million. In addition, the model has achieved significantly higher accuracy, particularly on the RSSCN7 and MSRSI datasets.

\subsubsection{Ablation Study}
\begin{table*}[h]
    \centering
    \caption{Comparison with different methods.}
    \begin{tabular}{|p{1.4cm}|p{5.6cm}|p{1.2cm}|p{1.2cm}|p{1.2cm}|p{1.2cm}|p{1.2cm}|p{1.2cm}|}
    \hline
        \textbf{Model Type} & \textbf{Model Structure} & \textbf{No of ViTs} & \textbf{No of CNNs} & \textbf{UCM} & \textbf{RSS-CN7} & \textbf{MS-RSI} & \textbf{Epo-chs} \\ \hline
        Fusion & ViT Base + Xception + ResNet152v2 + MobileNetV3Large  & 1 & 3 & 95.95 & 92.14 & 86.83 & 100 \\ \hline
        Fusion & ViT Base +  Xception + ResNet152v2 + MobileNetV3Large + EffecientNetB7 & 1 & 4 & 94.29 & 92.57 & 84.70 & 100 \\ \hline
        Fusion & ViT Base + Xception + ResNet152v2 + MobileNetV3Large + EffecientNetB7 + DenseNet121 & 1 & 5 & 93.86 & 92.68 & 88.27 & 100 \\ \hline
        Fusion & ViT Base +  InceptionResNetV2 + DenseNet121 + EffecientNetB7 + Xception + ResNet152v2 + MobileNetV3Large & 1 & 6 & 96.67 & 92.86 & 87.78& 100 \\ \hline
        Fusion & ViT Base +ViT Large + Xception + ResNet152v2 + MobileNetV3Large& 2 & 3 & 97.86 & 91.43 & 87.55 & 100 \\ \hline
        Fusion & ViT Base + ViT Large + Xception + ResNet152v2 + MobileNetV3Large + EffecientNetB7 & 2 & 4 & 96.43 & 92.14 & 87.20 & 100 \\ \hline
    Fusion & ViT Base +ViT Large + DenseNet121 + Xception + ResNet152v2 + MobileNetV3Large + EffecientNetB7 & 2 & 5 & 97.86 & 91.43 & 87.77 & 100 \\ \hline
    Fusion & ViT Base  + ViT Large + InceptionResNetV2 + DenseNet121 + EffecientNetB7 + Xception + ResNet152v2 + MobileNetV3Large & 2 & 6 & 97.38 & 93.75 & 86.73 & 100 \\ \hline
    Fusion + Soft Voting & (ViT Base + DenseNet121), (ViT Base + ResNet152v2) & 2 & 2 & 98.10 & 92.86 & 94.87 & 40 \\ \hline
    \textbf{Fusion + Soft Voting} & \textbf{(ViT Base+DenseNet121), (ViT Base+ResNet-152v2), (ViT Base+InceptionResNetV2), (ViT Base+Xception)} & \textbf{4} & \textbf{4} & \textbf{98.10} & \textbf{94.46} & \textbf{95.45} & \textbf{80} \\ \hline
    \end{tabular}
    \label{tab:compare}
\end{table*}

On three datasets, UCM, RSSCN7 and MSRSI, Table \ref{tab:compare} presents a comparison assessment of the performance of ten different model architectures. The accuracy of the models, which include different Convolutional Neural Networks (CNNs) and Vision Transformers (ViTs), is assessed after a specified number of training epochs. A distinct combination of ViTs and CNNs where the corresponding number of ViTs ranges from 1 to 2 and the number of CNNs ranges from 3 to 6. The models are optimized for fusion-based learning and utilize the improved characteristics of both ViTs and CNNs. On the UCM dataset, the model with 2 ViTs and 5 CNNs obtained the highest accuracy (97.86\%), whereas the model with 1 ViT and 5 CNNs had the lowest accuracy (93.86\%). On the RSSCN7 dataset, however, accuracy frequently falls lower, with the model with 2 ViTs and 6 CNNs achieving the highest accuracy of 93.75\%.  On the MSRSI dataset, the top-performing fusion model (1 ViT and 5 CNNs) achieved an accuracy of 88.27\%, while the model with 2 ViTs and 6 CNNs slightly underperformed at 86.73\%. The models incorporating one ViT had an average accuracy of 95.19\%, 92.56\%, and 86.90\% on the UCM RSSCN7, and MSRSI datasets, respectively. Notably, employing two ViTs within the models led to average accuracies of 97.63\%, 92.19\%, and 87.31\% on the corresponding datasets. The slight accuracy improvement with two ViTs on UCM and MSRSI but a decrease on RSSCN7 suggests that while multiple ViTs can enhance performance, they may also introduce redundancy or overfitting. In contrast, the models incorporating soft voting had a higher performance than the average of both the one and two ViT models. Additionally, they consumed fewer training resources than other models. All the fusion models performed better than single model classifiers since they can learn from diverse features coming from ViTs and CNNs. Additionally, achieving an accuracy of 98.10\% on UCM, 94.46\% on RSSCN7 and 95.45 \% on MSRSI, the application of soft voting fusion for specific combinations of ViTs and CNNs showed potential.

\begin{figure*}[h]
\centering
\begin{subfigure}{0.49\textwidth}
    \includegraphics[width=\textwidth,height=4.5cm]{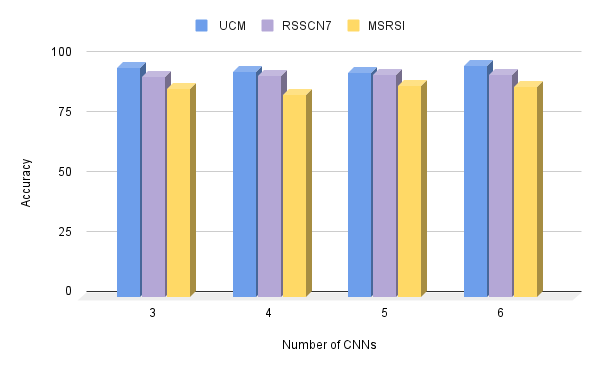}
    \caption{Accuracy over changing the number of CNNs with 1 ViT.}
    \label{fig:1vit}
\end{subfigure}
\hfill
\begin{subfigure}{0.49\textwidth}
    \includegraphics[width=\textwidth,height=4.5cm]{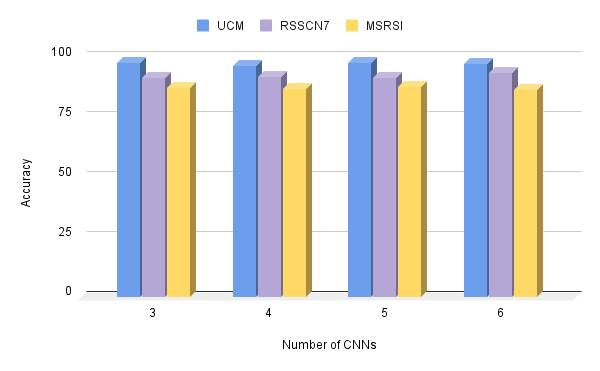}
    \caption{Accuracy over changing the number of CNNs with 2 ViT.}
    \label{fig:2vit}
\end{subfigure}
\caption{Accuracy over changing number of CNNs and ViTs.}
\label{fig:ablation}
\end{figure*}

As shown in Figure \ref{fig:ablation}, it has been noted that the number of ViT and CNN feature extractors has little effect on the overall performance of the fusion model. This phenomenon can be explained by the fact that, after a certain threshold, the extracted features from ViT and CNN tend to overlap as the number of feature extractors rises. The fusion model's accuracy consequently reaches saturation. Given that ViT and CNN employ various methods to capture meaningful visual data representations, there is a feature overlap between the two systems. While CNN uses convolutional layers to identify local patterns and structures, ViT uses self-attention mechanisms to extract global features from images. The fusion model benefits from a variety of representations obtained from both ViT and CNN in the beginning as the number of feature extractors rises. Beyond a certain point, however, the additional feature extractors begin to extract redundant or extremely similar data to the ones that already exist.

On the other hand, the problem of feature overlapping is resolved when a soft voting scheme is used in the fusion model. Soft voting is a method for combining predictions from different classifiers by giving each prediction a weight and calculating a weighted average. There is no direct overlap of features because soft voting takes each classifier's individual predictions into account separately. Soft voting's ability to accommodate a wider range of features that are complementary to one another is a key feature that contributed to the higher performance.

\subsection{Comparison with existing works}
We have evaluated our proposed solution with some existing literature. A direct comparison of UCM and RSSCN7 datasets is presented in Table \ref{tab:ucm-comp} and \ref{tab:comp-rsscn7}, respectively. Notably, since the MSRSI dataset is newly released, no direct comparisons were found. Since both UCM and RSSCN7 datasets are moderately balanced, the researchers relied on accuracy as the primary matrix. The majority of the research works have employed transfer learning for the classification of remote sensing images. On the UCM dataset, Thepade et al. \cite{thepade2022fusing} presented a deep fusion model that is further fine-tuned with machine learning classifiers. The extensive ablation study presents that InceptionV3 with ExtraTree classifier performs the best. Most of the research works, however, have fine-tuned single feature extractors. Due to the dependence on one feature extractor, some of the models have produced relatively low accuracy. 

\begin{table}[h]
    \centering
    \caption{Comparison of different models on UCM dataset.}
    \begin{tabular}{|p{4.5cm}|l|l|l|l|}
    \hline
        \textbf{Model} & \textbf{Accuracy} & \textbf{Precision} & \textbf{Recall} & \textbf{F1-Score} \\ \hline
         Snapshot ensemble CNN, dropCycle \cite{noppitak2022dropcyclic}& 97.38 & - & - & - \\ \hline
        Residual Network-based Features Extraction, Self-adaptive Global Best Harmony Search \cite{rajagopal2020fine}& 97.92 & - & - & - \\ \hline
        InceptionV3, Thepade’s Sorted Block Truncation Coding, ExtraTree \cite{thepade2022fusing}& 98.33 & 98.42 & 98.33 & 98.32 \\ \hline
        Custom Shallow CNN \cite{alem2022deep} & 88.1 & 89.9 & 88 & 89 \\ \hline 
        ResNet50 with Channel Attention \cite{zhuo2021remote}& 98.68 & - & - & - \\ \hline
        Multi-level Convolutional Pyramid Semantic Fusion \cite{sun2021multi} & 97.54 & - & - & - \\ \hline
        Proposed & 98.10 & 98.31 & 98.10 & 98.11 \\ \hline
    \end{tabular}
    \label{tab:ucm-comp}
\end{table}

On the RSSCN7 dataset, however, some researchers have experimented with other approaches. Zhang et al. \cite{zhang2020pruning} proposed a compressed network by pruning the unnecessary layers in CNN with an attention mechanism. The compressed network, however, achieved an accuracy of 83.03\%. Alhichri et al. \cite{alhichri2023rs} fine-tuned several pretrained models for remote sensing image classification. Among them, VGG16 achieved an acceptable 90.87\% accuracy. Following the comparison, the proposed model achieved a superior performance over similar approaches. 
\begin{table}[h]
    \centering
    \caption{Comparison of  different models on RSSCN7 dataset.}
    \begin{tabular}{|p{4.5cm}|l|l|l|l|}
    \hline
        \textbf{Model} & \textbf{Accuracy} & \textbf{Precision} & \textbf{Recall} & \textbf{F1-Score} \\ \hline
        Node Entropy \cite{yang2022remote} & 89.39 & - & - & - \\ \hline 
        Representation Learning, Resnet 50 \cite{li2021geographical} & 91.52 & - & - & - \\ \hline 
        Deep Color Fusion \cite{anwer2021compact} & 92.9 & - & - & - \\ \hline
        Filter Pruning With attention mechanism \cite{zhang2020pruning} & 83.93 & - & - & - \\ \hline 
        VGG16 \cite{alhichri2023rs} & 90.87 & - & - & - \\ \hline 
        Visual representation cross entropy \cite{zhang2023privacy} & 94 & 95 & 93 & 93 \\ \hline
        Proposed & 94.46 & 94.61 & 94.46 & 94.48 \\ \hline
    \end{tabular}
    \label{tab:comp-rsscn7}
\end{table}

Since both datasets are moderately balanced, the majority of the research works primarily focused on accuracy as the key evaluation metric. The direct comparison states that the model has the potential to achieve state-of-the-art classification performance. The superiority of the research work lies in the fusion of global and local features which is a crucial part of remote sensing image classification. Additionally, the performance saturation problem was addressed by incorporating the soft voting mechanism. While most research works achieve noteworthy accuracy, they often overlook the efficiency of training resources. Moreover, they mostly neglect thorough error analysis and the integration of Explainable AI techniques, which leaves room for potential biases. This research has carefully addressed both limitations, making a substantial contribution to the remote sensing image classification field.

\subsection{Explainability and Error Analysis}
\begin{figure*}
     \centering
     \begin{subfigure}[b]{0.32\textwidth}
         \centering
         \includegraphics[width=\textwidth,height=11cm]{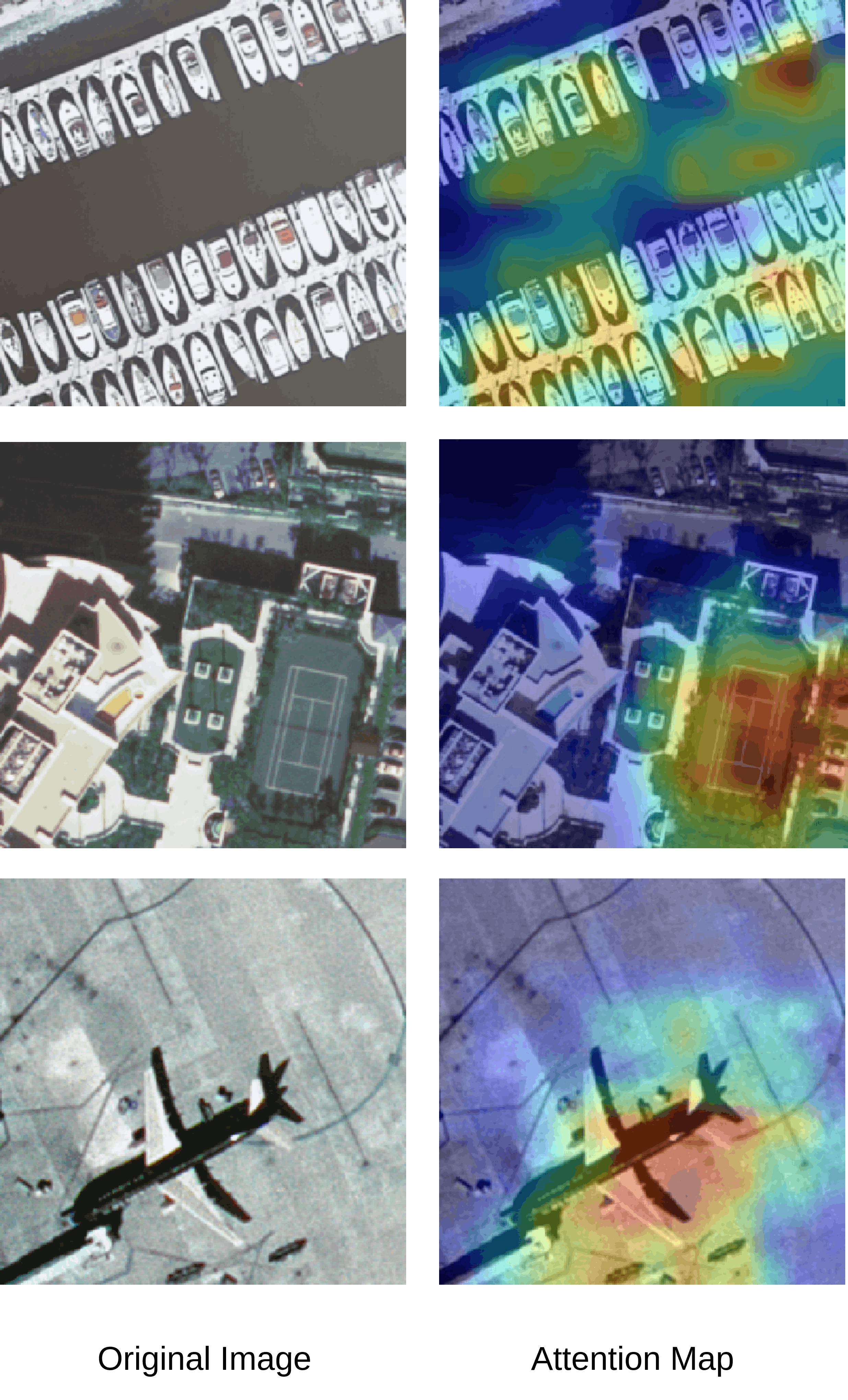}
         \caption{UCM dataset.}
         \label{fig:ucm_xai}
     \end{subfigure}
     \hfill
     \begin{subfigure}[b]{0.32\textwidth}
         \centering
         \includegraphics[width=\textwidth,height=11cm]{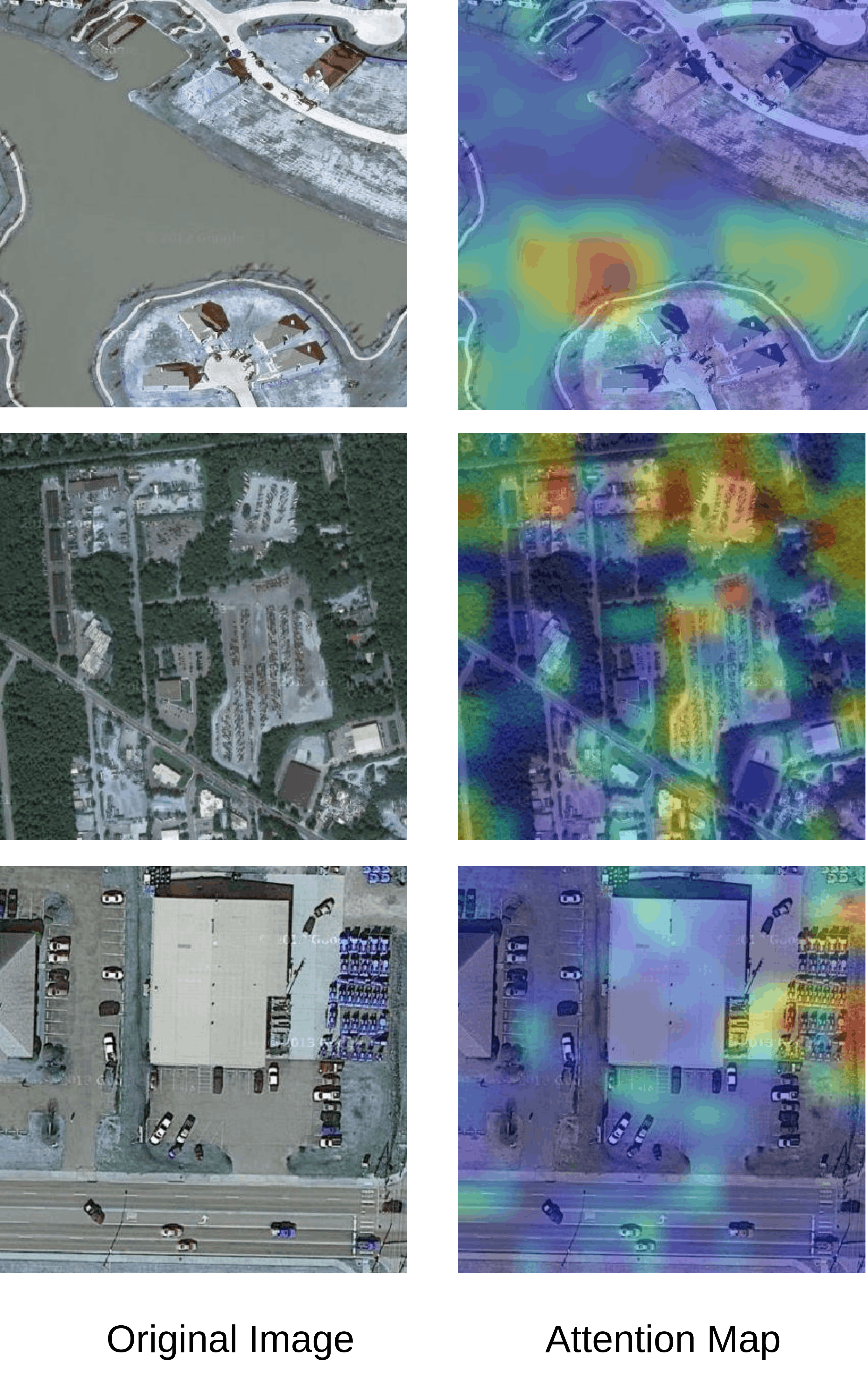}
         \caption{RSSCN7 dataset.}
         \label{fig:rsscn7_xai}
     \end{subfigure}
     \hfill
     \begin{subfigure}[b]{0.32\textwidth}
         \centering
         \includegraphics[width=\textwidth,height=11cm]{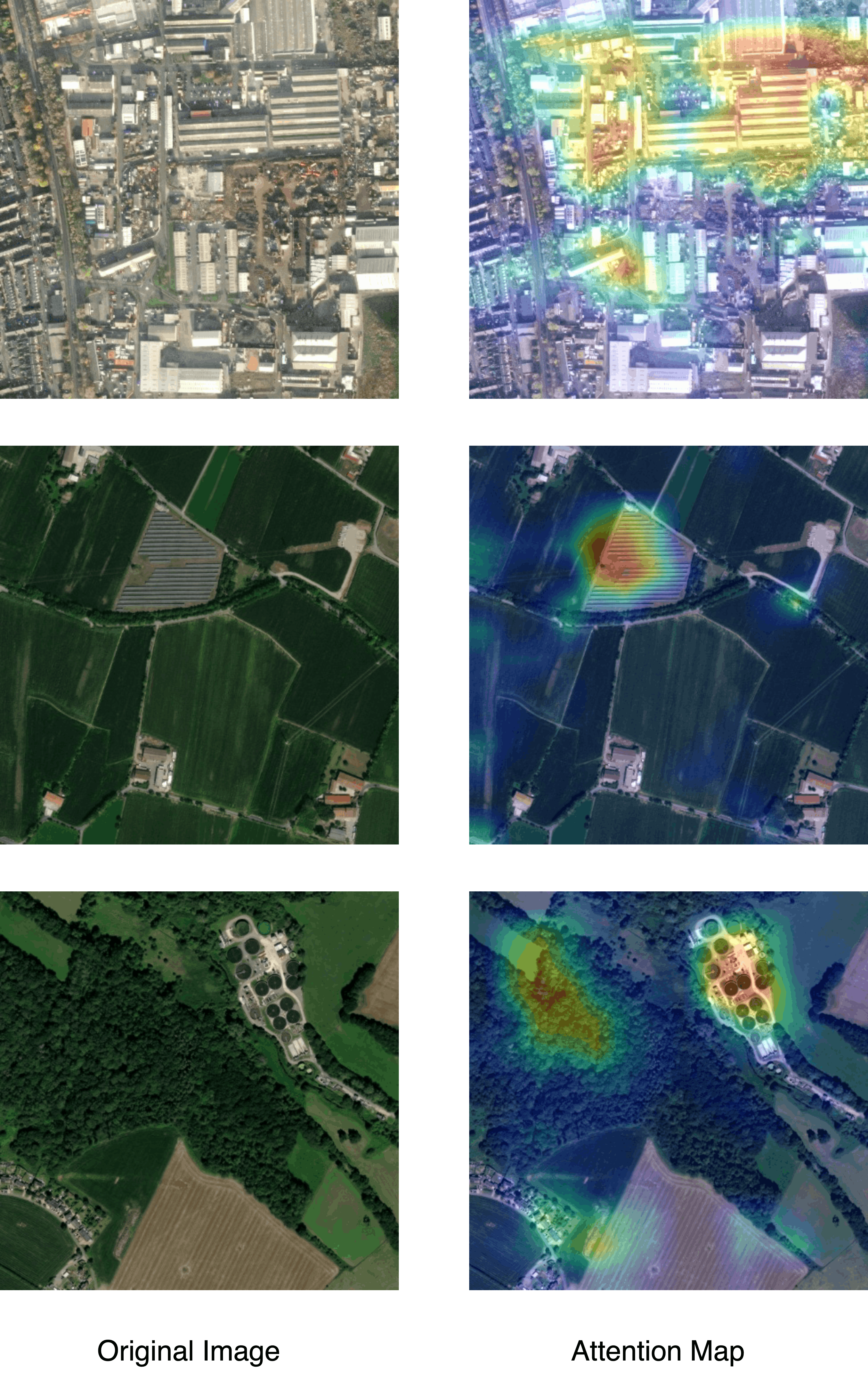}
         \caption{MSRSI dataset.}
         \label{fig:MSRSI_xai}
     \end{subfigure}
     \hfill
\caption{Attention maps of a fusion model on different datasets.}
\label{fig:xai}     
\end{figure*}

In view of the comparison presented in Table \ref{tab:ucm-comp} and \ref{tab:comp-rsscn7}, the proposed system achieves a competitive accuracy. However, there still remains a need for better model interpretability and bias detection. To understand the model's decision-making process, we have presented the model's attention maps. For producing the attention maps, we have leveraged Gradient-weighted Class Activation Mapping (Grad-CAM) \cite{selvaraju2017grad}. Figure \ref{fig:ucm_xai} holds the attention heatmap of a fusion model (ViT Base + ResNet152V2) on the UCM dataset. The figure shows that the model has attained the correct regions of the images for making the perfect decision. Likewise, Figure \ref{fig:rsscn7_xai} presents the attention heatmap on the RSSCN7 dataset of the same classifier. The examples illustrate that the model prioritizes water features when examining images of rivers. Similarly, the building and the cars are focused on the residents and the parking classes respectively. Likewise, Figure \ref{fig:MSRSI_xai} illustrates the attention map on the MSRSI dataset. Similar to the previous examples, the model has correctly focused on the key areas, particularly in the images of the solar panel and wastewater plant.

\begin{figure*}
    \centering
    \includegraphics[width=\textwidth,height=6cm]{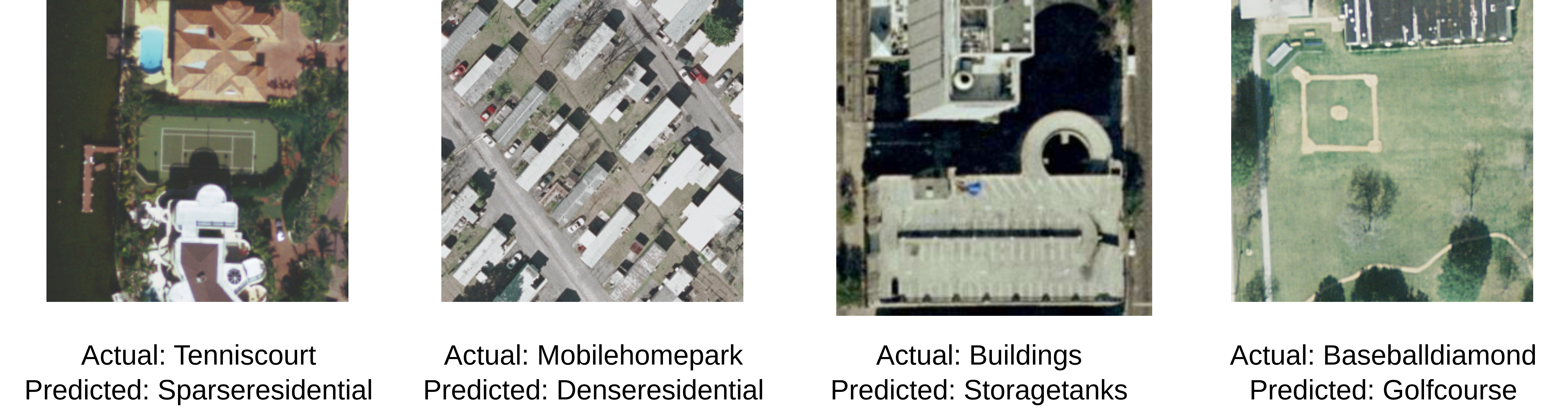}
    \caption{Error analysis of the proposed model on the UCM dataset.}
    \label{fig:ucm_errors}
\end{figure*}

\begin{figure*}
    \centering
    \includegraphics[width=\textwidth,height=6cm]{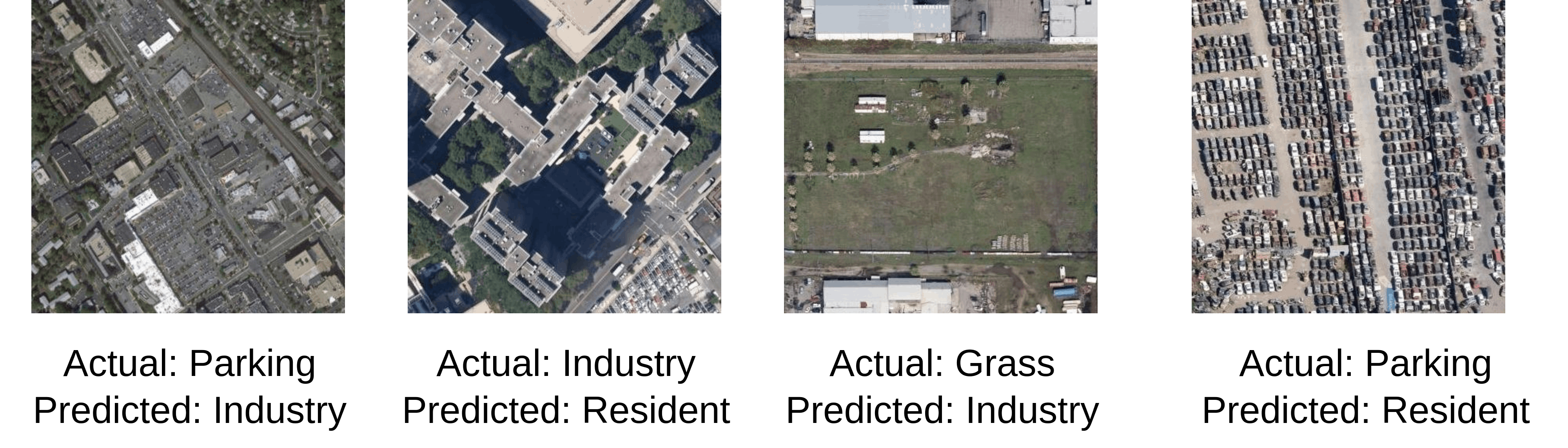}
    \caption{Error analysis of the proposed model on the RSSCN7 dataset.}
    \label{fig:rsscn7_errors}
\end{figure*}

\begin{figure*}
    \centering
    \includegraphics[width=\textwidth,height=6cm]{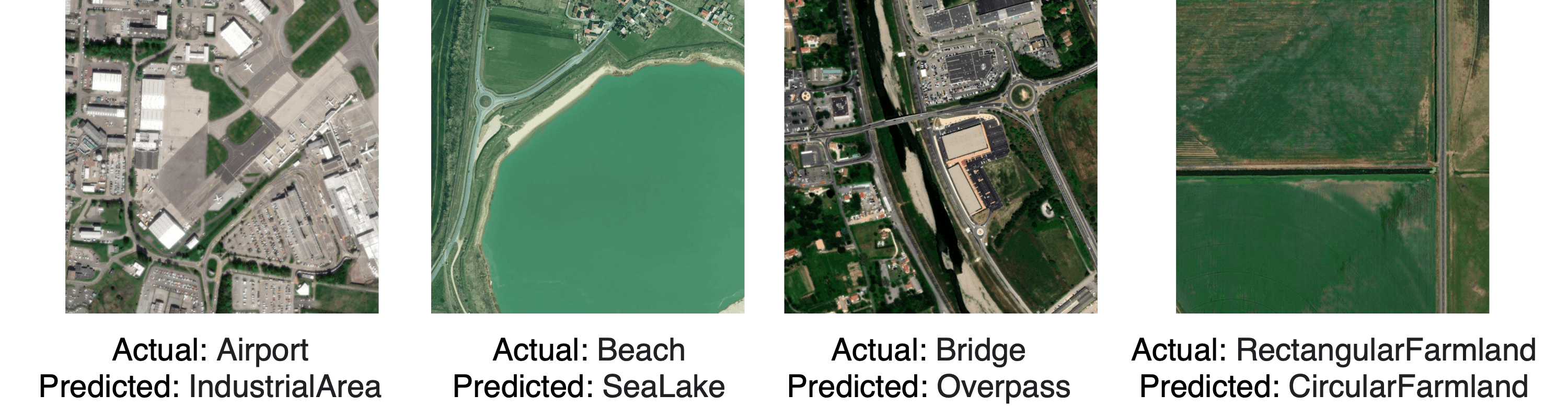}
    \caption{Error analysis of the proposed model on the MSRSI dataset.}
    \label{fig:msrsi_errors}
\end{figure*}

The proposed model, however, makes some incorrect predictions. Figure \ref{fig:ucm_errors}, \ref{fig:rsscn7_errors},  \ref{fig:msrsi_errors} holds some incorrect predictions of the model on the UCM, RSSCN7, and MSRSI datasets, respectively. The errors are primarily due to high interclass similarity and little focus on local features. According to the analysis, the model successfully captured global features but struggled with fine-grained details in certain cases. Thus, the global feature became dominant and adversely influenced the classification.

\subsection{Discussion}
We have developed an ensemble model combining two-stream fusion models, where each model has classification ability. The architecture of the model is carefully designed to be resource-efficient while training. A comparative study demonstrates the superior performance of the proposed solution. Notably, the proposed solution leverages existing techniques like CNN and ViT. Nonetheless, the major contribution of this research lies in the effective integration of these modules. Our study shows that combining ViT and CNNs leads to performance saturation due to overlapping feature extraction, which can be mitigated through soft voting. Additionally, we have shown that instead of training large models with a high number of epochs, training smaller models with fewer epochs and combining their predictions through a voting mechanism is an effective solution that elevates the performance and reduces resource consumption. However, due to the integration of four fusion models, the model consumes relatively high memory while testing. Nevertheless, since the four models have independent classification abilities, the inference process can be parallelized by running each fusion model separately. Furthermore, the models can be compressed using different techniques, including quantization and pruning. Since compression techniques can potentially degrade performance, a comprehensive study is needed to ensure that performance is maintained while minimizing resource consumption. Lastly, although the architecture is primarily designed for classifying remote sensing images, its strong discriminative capability makes it suitable for other applications as well—such as remote sensing image retrieval, with minimal modifications. To enable retrieval, the class label can be encoded and fused with the extracted image features, resulting in a compact and informative representation of the remote sensing data as demonstrated in previous works \cite{kang2020deep,sun2021multisensor,sun2022unsupervised}. Overall, the model can be a reliable option for remote sensing image classification, considering the performance and the computational efficiency of training.

\section{Conclusion}
\label{sec:conclusion}
In this research, a novel and efficient architecture is presented for classifying the remote sensing images. The introducing model combines CNN architecture with Vision Transformers (ViTs), and in order to increase the prediction probability, soft voting is also used. The proposed fusion model successfully capitalizes on the expertise of both CNNs and ViTs, as well as the benefits of the soft voting mechanism to produce a robust classifier. The UC Merced Land Use (UCM), RSSCN7, and MSRSI datasets were used to thoroughly evaluate the model. The dataset is balanced using geometric augmentation, and the gamma transformation is applied to enhance the image quality. The results obtained for thee datasets, 98.10\%, 94.46\%, and 95.45\%, respectively, demonstrate outstanding efficacy in identifying different types of remote sensing images.  The accuracy attained by the model demonstrates its ability to handle diversified land cover classes and remote sensing images. A comparison assessment with different models is performed and it is observed that the proposed fusion model outperforms other architectures substantially, showing its superiority in classification tasks. The evaluation metrics also further validate the model's stability and reliability. This system, however, consumes a relatively high amount of memory, which can be compressed using quantization techniques. Nonetheless, these techniques may lead to a reduction in classification accuracy. In the future, a study could be conducted to reduce memory consumption without compromising performance. In conclusion, the suggested fusion model represents a substantial development in classifying remote sensing images. It establishes new opportunities for utilizing deep learning's potential in various remote sensing applications. The fusion of these two architectures offers a comprehensive solution to address the challenges of classifying a large-scale image and contributes to the exploration and knowledge extraction from remote sensing data.

\section*{Supplementary Materials}
The experiment can be found at this link: \url{https://github.com/NifulIslam/Remote-Sensing-Image-Classification-With-ViT-and-CNN/tree/main}. The readme file presents the detailed methodology at a glance.

\bibliographystyle{unsrt}
\bibliography{cites}

\end{document}